\documentclass{article}

    \usepackage[preprint]{neurips_2019}

\usepackage[utf8]{inputenc} 
\usepackage[T1]{fontenc}    
\usepackage{url}            
\usepackage{booktabs}       
\usepackage{amsfonts}       
\usepackage{nicefrac}       
\usepackage{microtype}      

\usepackage{soul}
\usepackage{url}
\usepackage[hidelinks]{hyperref}
\usepackage[utf8]{inputenc}
\usepackage[small]{caption}
\usepackage{graphicx}
\usepackage{amsmath}
\usepackage{algorithm}
\usepackage{algorithmic}
\usepackage{amsmath,amssymb,amsfonts}

\usepackage{caption}
\usepackage{subcaption}
\usepackage{placeins}

\title{Target-Focused Feature Selection Using a Bayesian Approach}

\author{%
  Orpaz Goldstein \\
   Department of Computer Science\\
   University of California, Los Angeles\\
  \texttt{orpgol@cs.ucla.edu} \\
   \And
   Mohammad Kachuee \\
   Department of Computer Science\\
   University of California, Los Angeles\\
   \texttt{mkachuee@cs.ucla.edu} \\
   \AND
   Kimmo Karkkainen \\
   Department of Computer Science\\
   University of California, Los Angeles\\
   \texttt{kimmo@cs.ucla.edu} \\
   \And
   Majid Sarrafzadeh \\
   Department of Computer Science\\
   University of California, Los Angeles\\
   \texttt{majid@cs.ucla.edu} \\
}

\begin{document}
\maketitle

\begin{abstract}
In many real-world scenarios where data is high dimensional, test time acquisition of features is a non-trivial task due to costs associated with feature acquisition and evaluating feature value. The need for highly confident models with an extremely frugal acquisition of features can be addressed by allowing a feature selection method to become target aware.
We introduce an approach to feature selection that is based on Bayesian learning, allowing us to report target-specific levels of uncertainty, false positive, and false negative rates. In addition, measuring uncertainty lifts the restriction on feature selection being target agnostic, allowing for feature acquisition based on a single target of focus out of many. We show that acquiring features for a specific target is at least as good as common linear feature selection approaches for small non-sparse datasets, and surpasses these when faced with real-world healthcare data that is larger in scale and in sparseness.
\end{abstract}

\section{Introduction}
As big data becomes ubiquitous so does the increase in the dimensionality of data. As the selection of features increases, feature selection becomes a necessary tool in the evaluation and acquisition of features \cite{li2018feature}, and in turn for the training of learning models. This is increasingly true in the healthcare domain, where data is accumulated and under-utilized ~\cite{groves2013big,wang2018big}. Moreover, in the healthcare domain, both budget for features and model uncertainty should be taken into account in order for a feature selection model to be practical. Since in many cases our main target of interest is the minority target, we would rather focus on reducing the uncertainty of a specific target of interest rather than the general uncertainty, while maintaining a budget for features. For example, two types of heart disease might display similar symptoms, but we rather focus our resources understanding whether the patient has a less common disease which is more fatal.

Classic approaches to feature selection focus on maximizing information gain and inferring feature relevance \cite{bentz1997selection,guyon2003introduction}. 
Health informatics methods of feature selection take into account real-world costs associated with the acquisition of features and the need to maintain a budget. Costs of tests, physician time, patient discomfort, should all be taken into account when reasoning on which feature is to be acquired, by using cost-sensitive decision methods or active sensing \cite{freitas2007cost,yu2009active}. In addition to costs, changes in medical data availability might call for iterative feature aggregation in training time, requiring an online cost-sensitive budgeted approach \cite{kachuee2018opportunistic}.

Healthcare data tends to be imbalanced, some conditions or variants of a disease are more common. Some targets carry more significance or are more relevant to a specific diagnosis. Acquisition of relevant data is made possible using an active learning approach \cite{ijcai2018-486}. Contributing to the imbalance is also the sparseness of data. Due to the high dimensionality of the data, not all data points will have all features. For medical domain feature selection and prediction, ensemble methods have been used to reduce the effects of imbalance data, and of inherent missingness \cite{huda2016hybrid,liu2006healthcare}, and more recently with a robust feature selection framework \cite{ijcai2018-443}. While addressing the imbalance in data is closely related to our work, the acquisition of features that are germane to a specific target of focus is not addressed.

Uncertainty measurement in a machine learning model flows from applying a probabilistic approach to learning, also known as Bayesian learning. Sampling a trained probabilistic model for latent variables allows us to capture the inherent uncertainty in the model.  The usage of Gaussian weight distributions to estimate the uncertainty was first discussed in \cite{denker-lecun-91}. Later work includes \cite{Bishop:2006:PRM:1162264,Murphy:2012:MLP:2380985,pmlaai} and many more.
Application of uncertainty to feature selection robustness appears in Same Decision Probability (SDP) \cite{Choi2012SamedecisionPA}, which measures the effect of feature acquisition on the shift in the decision making of a model.  SDP measures the uncertainty in the model while acquiring features, and reasons on stopping criteria based on a threshold of confidence and budget. More recently, an expected SDP query and an optimal feature selection algorithm based on SDP were proposed \cite{choi2017optimal}. SDP queries are generally $PP^{PP}\textit{-complete}$ which makes it costly for many high dimensional real-world applications.

In this paper, we propose a novel probabilistic uncertainty-based method, for target-specific feature acquisition. 
Our first contribution is providing a method to focus resources on a single target of interest, that is generalizable, scalable, and consistent in selecting informative features for a specific single target of interest out of many.

Our second contribution is polynomial time, threshold-based method, allowing us to reason on model confidence in predictions while learning a representation of the data, and make a decision on whether to ask for more data or declare readiness to start making predictions on real-world scenarios.

\section{Background}
In order to capture uncertainty in a model, we need to learn a representation of a latent distribution over a set of parameters defining that distribution and be able to sample the learned parameters in order to associate the captured uncertainty with test time examples of the data and targets. Our optimization function, therefore, will be taking a probabilistic approach.

Using Variational Inference we will estimate $\lambda^\star$ using Kullback-Leibler (KL) divergence such that:
\begin{equation}\label{eq:1}
\lambda^\star=\textit{arg min}_\lambda\textit{KL} (q(\mathbf{z};\lambda) \ ||\  p(\mathbf{z}|\mathbf{X})),
\end{equation}
where $q(\mathbf{z};\lambda)$ is the estimation of posterior distribution $p(\mathbf{z}|\mathbf{X})$ optimized over parameters $\lambda$.

Since the posterior $p(\mathbf{z}|\mathbf{X})$ is unknown to us, we will resort to maximizing the Evidence Lower Bound (ELBO) as an optimization function:
\begin{equation}\label{eq:2}
\textit{ELBO}(\lambda)=\mathrm{E}_{q(\mathbf{z};\lambda)}[\textit{log}\ p(\mathbf{X},\mathbf{z})-\textit{log}\ q(\mathbf{z};\lambda)], 
\end{equation}
which is equivalent to minimizing KL divergence \cite{Jordan1999,Bishop:2006:PRM:1162264,DBLP:journals/corr/KingmaW13}.

Gradient optimization of ELBO is done via the reparameterization trick \cite{DBLP:journals/corr/KingmaW13}.
\begin{equation}\label{eq:4}
\resizebox{.7\linewidth}{!}{$
    \displaystyle
\nabla_\lambda\textit{ELBO}(\lambda)\approx\frac{1}{s}\sum^{s}_{s=1}[\nabla_\lambda(\textit{log}\ p(\mathbf{X},\mathbf{z}(\epsilon;\lambda))-\textit{log}\ q(\mathbf{z}(\epsilon;\lambda);\lambda))],
$}
\end{equation}
where $s$ is the number of samples drawn.

\section{Target-Focused Feature Selection}

\subsection{Problem Set Up}
Using a minimal amount of features, our goal is to achieve reasonable confidence for a specific class, as described in our objective function.
\begin{align}
        \textit{argmax}_{\textit{FS}} \ (\textit{confidence}_\theta - \sum_{f_i \in \textit{FS}} \frac{1}{v_i}),
\end{align}\label{eq:0}
\begin{align*}
    \textit{Subject to:\;\;} |\textit{FS}| < \beta.
\end{align*}

Such that $\textit{FS}$ is the set of acquired features we wish to minimize, $|\textit{FS}|$ is the cardinality of the set, $v_i$ is the value associated with each feature. The objective is to frugally acquire the most valuable features while achieving maximum confidence in a specific class $\theta$, without exhausting our budget for features $\beta$.

\subsection{Feature Value Measurement and Acquisitions} 
Evaluation of features per target considers the contribution of each feature towards minimizing uncertainty for our target of interest, jointly evaluated with the features already acquired. In addition to confidence scores, features vectors are scored for their cosine similarity as well as their Hamming weight scores, in order to gauge potential information gain from a candidate feature.

In order to use $\textit{ELBO}$ as our optimization function, we model the linear regression case in which our $\lambda$ contains the input $\mathbf{X}$, a single layer of weights $\mathbf{W}$ and a bias $\mathbf{b}$ such that $\lambda = (\mathbf{W},\mathbf{b},\mathbf{X})$. Here $\mathbf{X}\in\mathbb{R}^{c,r}, w\in\mathbb{R}^{c,d}, b\in\mathbb{R}^{1,d}$. $\mathbf{X}$ has $r$ data points and $c$ features, and the model will learn the distribution over $d$ targets. Assuming independence given our parameters:
\begin{equation}\label{eq:5}
  p(\mathbf{z}|\mathbf{W},\mathbf{b},\mathbf{X}) = \prod_{n=1}^{r}p(\mathbf{z}_n|\mathbf{X}_n^\top \mathbf{W}+\mathbf{b},\sigma^2_z),
\end{equation}
where $z$ is the $\textit{ELBO}$ optimized posterior estimation. 
We define the priors on both parameters to be the standard normal distribution.

\subsubsection{Algorithm Overview}
For each feature not already in our feature set $f_i \notin\textit{FS}$, a model estimating $\textit{ELBO}(\lambda)$ is trained for each $f_i \cup \textit{FS}$. Once trained, each feature is scored on its contribution to model confidence in predicting a specific target on a validation set, in addition to the cosine similarity and co-variance scores between the feature $f_i$ and all the features already in $\textit{FS}$. We then select a single feature $f_i \notin \textit{FS}$, to be aggregated together with the features already in $\textit{FS}$, based on the scores received in the previous step. With each feature added, a new model is trained trying to estimate the latent target variables on a previously unseen test set. We continue aggregating features until we reach a stopping condition or we exhaust our budget as described in Algorithm~\ref{alg:algorithm}. Our algorithm runs in $O(n^2)$ time. Complete time complexity analysis provided in supplemental material.

\begin{algorithm}[t]
\caption{Target-Focused Feature Selection}
\label{alg:algorithm}
\textbf{Input}: $\beta:\textit{Budget};$ $\quad~\textit{FPT}, ~\textit{FNT}, ~\textit{CT}:\textit{Thresholds}$; \\
$\mathbf{X}:\textit{Train set}$; ~$\mathbf{X}^{'}:\textit{Validation set}$;  ~$\mathbf{X}_{test}:\textit{Test set}$; \\
$\mathbf{y}: \textit{Targets for} ~\mathbf{X} ,\mathbf{X}^{'} \textit{and} ~\mathbf{X}_{test}$; \\
$\mathbf{F}: \{f_1,f_2,...,f_n\}, \textit{set of available features}$ \\
\textbf{Parameter}: $\textit{FS} \leftarrow \{\}: \textit{features selected}$; \\
$M \leftarrow \textit{model optimizing ELBO}(\lambda)$ \\
$F \leftarrow \textit{function for computing} ~v_i:~\textit{value for feature i}$;\\
$\textit{FP}_\theta$, $\textit{FN}_\theta$, $\textit{confidence}_\theta:\textit{current false positive,}$\\ $\textit{false negative and confidence for specific target }\theta$\\
\textbf{Output}: $\textit{SF} \subseteq \mathbf{F} ~\textit{within budget} ~\beta$
\begin{algorithmic}[1] 
\WHILE{$|\textit{SF}| < \beta, \textit{FP}_\theta > \textit{FPT}, \textit{FN}_\theta > \textit{FNT},$ \\ $\textit{confidence}_\theta < \textit{CT}$ ~}
\FOR{$f_i \in \mathbf{F}$}
\STATE $\textit{Train} ~M(\mathbf{X}, ~\textit{SF} \cup f_i, ~y)$
\STATE $v_i \leftarrow F(~M(\mathbf{X^{'}}, ~\textit{SF} \cup f_i, ~y), ~\textit{FS}, f_i)$
\ENDFOR
\STATE $\textit{FS} \leftarrow \textit{FS} \cup f_i ~|~ \textit{argmax}_i ~ v_i \in V$
\STATE $\textit{FP}_\theta$, $\textit{FN}_\theta$, $\textit{confidence}_\theta \leftarrow ~M(~\mathbf{X}_{test}, ~\textit{FS}, ~y)$
\ENDWHILE
\STATE \textbf{return} solution
\end{algorithmic}
\end{algorithm}

\subsubsection{Measuring Per-Target-Uncertainty}
Our available data is split into a training set $\mathbf{X}_{\textit{train}}$ and a testing set $\mathbf{X}_{\textit{test}}$. To obtain our input $\mathbf{X}$ we sample the training data $\mathbf{X}_{\textit{train}}$ in a balanced way. For example, if we are trying to predict 3 targets then $\mathbf{X}$ will have $33\%$ of the data points correspond to each of our targets, regardless of the original distribution. In order to generate a validation input dataset $\mathbf{X}'$, we sample $\mathbf{X}_{\textit{train}}$ according to its original distribution (no balancing).

At each iteration, a subset of all available feature $f_i \cup \textit{FS}$ is trained to learn $\lambda = (\mathbf{W},\mathbf{b},\mathbf{X})$.
Once trained, we score the feature subset on the on the validation set $\mathbf{X}'$ by measuring the effect acquired features had on a per-target uncertainty. Using our learned distribution, we sample each of our parameters such that $\mathbf{W}' \sim \mathbf{W}, \mathbf{b}' \sim \mathbf{b}$ and calculate the probability vector: 
\begin{equation}\label{eq:6}
	\textit{prob} = \mathbf{X}'^{\top}\mathbf{W}'+\mathbf{b}',
\end{equation}
where $\textit{prob}\in\mathbb{R}^{r,d}$ has the probability of each data point belonging to each possible target. We then get the prediction vector by calculating softmax for each $\textit{prob}_i$:
\begin{equation}\label{eq:7}
	\mathbf{y}_i = \textit{argmax (softmax}(\textit{prob}_i)) = \frac{\textit{exp}(\textit{prob}_i)}{\sum_d\textit{exp}(\textit{prob}_{i,d})}.
\end{equation}
 Next we evaluate precision, represented by the fraction of times that $\mathbf{y}_{{\theta}}$ corresponding to target $\theta$, was equal the correct target for position $i$. Note that $\mathbf{y}_{\theta} \in \mathbf{y}$, and is of subset size $|\mathbf{y}_{\theta}|$:
\begin{equation}\label{eq:8}
	\textit{precision}_\theta = \frac{1}{|\mathbf{y}_{\theta}|}\sum_{i=1}^{|\mathbf{y}_{\theta}|}1(\mathbf{y}_{\theta,i} = \theta),
\end{equation}
where $1(\mathbf{y}_{\theta,i} = \theta)$ equals 1 if data point $\mathbf{y}_{\theta,i}$ has the target value $\theta$, and 0 otherwise.

Repeating \ref{eq:6} - \ref{eq:8} for $l$ iterations, sampling the distribution of our parameters each time, our confidence score becomes the averaged precision over multiple iterations. Therefore the confidence for a specific target:
\begin{equation}\label{eq:9}
	\textit{confidence}_\theta = \frac{1}{l} \sum_{j=1}^{l}(\textit{precision}_{\theta,j}).
\end{equation}
Here $l$ is the number of times we sample our learned distributions. The trade-off using $l$ is between a more accurate representation of the model confidence, and a faster model. We have found that 300 iterations were accurate enough in  reporting confidence in our case. 

\subsubsection{Adding Vector Similarity Scores}
In addition to the confidence scores, we wish to capture the potential information gain of the current candidate feature $f_i$ given the existing features in $\textit{FS}$. We use the computed similarity scores: co-variance distance score, and cosine similarity score. We sum the inverse scores for all such pairwise comparison and then normalize to the range [1,0].
\begin{equation}\label{eq:10}
    \textit{CovScore} =  N^{0,1}\sum_{g_i \in \textit{FS}} 1-\textit{cov} (g_i, f_i),
\end{equation}
\begin{equation}\label{eq:11}
    \textit{CosScore} = N^{0,1} \sum_{g_i \in \textit{FS}} 1-\textit{cos}(g_i, f_i).
\end{equation}
CovScore and CosScore are the summed inverse co-variance distances and cosine similarities, transferred to the [0,1] range applying the normalization $N^{0,1}$.

Our final feature value for the current feature $f_i$ is then
\begin{equation}\label{eq:10}
    v_i = \omega_1*\textit{confidence}_\theta + \omega_2*\textit{CovScore} + \omega_3*\textit{CosScore}
\end{equation},
where $\omega_1, \omega_2, \omega_3$ are hyperparameters.

Once all features $F_i \notin \textit{FS}$ have been scored and evaluated for their contribution towards class $\theta$ as part of set $\textit{FS}$, we append the single feature that maximized $v_i$ to the set $\textit{FS}$

\section{Evaluation}
Here we provide an empirical evaluation of our target focused method (TF) compared with prevalent linear feature selection techniques.

\textbf{Mutual Information} (MI) is estimating statistical dependency for feature selection \cite{kraskov2004estimating}, and is widely used as a non-parametric approach to evaluating data dependencies. The MI approach works by estimating correlation level based on entropy from k-nearest neighbor distances.

\textbf{Max-relevance min-redundancy} (mRMR) \cite{peng2005feature} is a first-order incremental feature selection method based on Mutual Information that eliminates redundancy in features while selecting relevant ones.

\textbf{Least absolute shrinkage and selection operator (Lasso) model} \cite{tibshirani1996regression} is an L1-based feature selection approach. Performing some regularization in addition to filtering out unwanted features, Lasso is an "automatic" approach to feature selection. 

\textbf{Extremely randomized trees (Extra trees)} \cite{geurts2006extremely} is a tree-based model performing feature selection based on the importance values computed by the model. 

We remind the reader that all methods mentioned above are target agnostic, and therefore we compare confidence in both the specific target of interest as well as the classic general confidence of a model (can be seen in the supplemental material) over all targets in the data.

\subsection{Datasets}
We evaluated our model on image classification task, as well as a breast cancer detection task, both chosen from the UCI machine learning repository \cite{Dua:2017}, in addition to various disease prediction tasks assembled using the Centers for Disease Control and Prevention's (CDC)  National Health and Nutrition Examination Survey (NHANES) \cite{nhanes} data. 

For each of our sets, we select a target of special interest, that we would like our model to focus on when deciding which features to acquire. Projecting this to the real world, the focus target will be a specific health issue in a dataset of symptoms and possible tests or images, pointing to more than one possible target class. 

The data is as follows: From the UCI machine learning repository, we use SatLog data \footnote{Available here: \href{https://archive.ics.uci.edu/ml/datasets/Statlog+(Landsat+Satellite)}{UCI Statlog (Landsat Satellite)}}. A dataset of evaluating image data and identifying a particular type of soil in satellite images. Also from UCI, we use the Breast Cancer Wisconsin dataset \footnote{Available here:  \href{https://archive.ics.uci.edu/ml/datasets/Breast+Cancer+Wisconsin+(Diagnostic)}{UCI Breast Cancer Wisconsin}}. Providing features that are computed from a digitized image of a fine needle aspirate (FNA) of a breast mass. From NHANES, we construct two datasets ourselves based on the approach described by \cite{kachuee2019nutrition}. One for evaluating diabetes, and one for evaluating heart diseases. 
To construct our datasets we join all possible NHANES tables that are correlated with our targets. For example, for the heart disease dataset, we join all tables that have features with correlation to any of 5 heart conditions. This causes the resulting sets to have a vast amount of possible features. 

Dataset statistics, as well as the target chosen for each dataset, is listed in Table \ref{data_stat}.
For the NHANES datasets, targets are renamed from the original data for convenience. Blood glucose refers to the feature LBXGLU, the amount of glucose in the blood when fasting, used here to indicate whether or not an individual has diabetes. Congestive heart failure (CHF) refers to the feature MCQ160B, and it is one of 5 heart conditions we construct the dataset for (MCQ160E, MCQ160F, MCQ160C,  MCQ160B, MCQ180B).

\subsection{Evaluation Methodology}
Assuming a constant budget for features, we run all feature selection approaches on the same training subset of the data and iteratively evaluate for each feature we add. We select a single target to act as the focus of our method. We put an emphasis on the model confidence for that specific target value that we wish to maximize over all targets. The target chosen for each dataset is listed in Table \ref{data_stat}.

The compared models were constructed with the following parameters:

(i) Mutual information (MI) between our training data and the training target was calculated using a different number of neighbors. Balancing the estimation variance and bias, we evaluated number of neighbors $k \in [1,2,3,5,10]$. The $k=3$ instance, giving the best average result in all cases was selected.

(ii) mRMR was evaluated both on "MIQ" and "MID" feature selection methods.

(iii) Lasso with cross-validation was used in this experiment. In order to find the best $\alpha$ value for the regularization process, we considered $\alpha \in [1,0.1,0.001,0.0005,0.0001]$. In addition, the best set up of Lasso for the average case was as follows: a maximum number of iterations was set to 1000, tolerance was set to 0.1, and the number of cross-validation folds set to 10.

(iv) Extra trees classifier was used in our experiments. The number of estimators in this model was set to 1000, with no maximum depth defined. In order to split a node, the minimum number of samples was set to 2, and the quality of split measured by Gini impurity.

(v) Our Target-Focused (TF) feature selection was trained using $\omega_1 = 0.4, \omega_2 = \omega_3 = 0.3$. 

The machine used for evaluation had specification: Intel 12 core i9-7920x (2.90GHz) CPU, 128 GB RAM, and 4 GeForce RTX 2080TI GPUs.

\section{Results}
In this section, we will report confidence, false positive, and false negative scores, as well as F1 scores of our model at intervals as features are acquired. When plotting model trends of the aforementioned metrics, we denote the variance scores of model confidence prior to applying the argmax function (equation \ref{eq:7}), as the line margin on confidence plots, as can be seen in plots below.

\begin{table}[t]
\centering
\resizebox{0.7\linewidth}{!}{
 \begin{tabular}{ c c c c c c } 
 \hline
 Dataset & Size & Features & Targets  & Focus target & Missingness \\ [0.5ex] 
 \hline
  UCI Breast cancer & 569 & 32 & 2 & Malignant & 0\% \\
  UCI Satlog & 4435 & 37 & 6 & Damp grey soil & 0\% \\
  NHANES Diabetes& 25474 & 581 & 2 & Blood Glucose & 25\% \\
  NHANES Heart& 49346 & 555 & 5 & CHF & 25\% \\
   \hline
  \end{tabular}
  }
  \caption{Datasets statistics}
\label{data_stat}
\end{table}

\begin{table*}[t]
\centering
\resizebox{0.85\textwidth}{!}{
 \begin{tabular}{ c | c c c c c c c c c c c c c c c c c c c c c c c c c c c c} 
 \toprule
  \multicolumn{21}{c}{\textbf{F1 scores}} \\
 \hline
 & \multicolumn{10}{c|}{\textbf{Breast Cancer}} & \multicolumn{10}{c}{\textbf{Satlog}} \\
  \hline
 \textbf{f} &\multicolumn{2}{c}{\textbf{MI}} &
 \multicolumn{2}{c}{\textbf{mRMR}} &
 \multicolumn{2}{c}{\textbf{Lasso}} & \multicolumn{2}{c}{\textbf{Extra Trees}} & \multicolumn{2}{c|}{\textbf{TF}}
 &\multicolumn{2}{c}{\textbf{MI}} &
 \multicolumn{2}{c}{\textbf{mRMR}} &
 \multicolumn{2}{c}{\textbf{Lasso}} & \multicolumn{2}{c}{\textbf{Extra Trees}} & \multicolumn{2}{c}{\textbf{TF}} \\
  \hline
 5 & \multicolumn{2}{c}{0.93} &
 \multicolumn{2}{c}{0.94} &
 \multicolumn{2}{c}{0.80} & \multicolumn{2}{c}{0.91} & \multicolumn{2}{c|}{0.95}
 &\multicolumn{2}{c}{0.70} &
 \multicolumn{2}{c}{0.27} &
 \multicolumn{2}{c}{0.35} & \multicolumn{2}{c}{0.08} & \multicolumn{2}{c}{0.86} \\
 10 & \multicolumn{2}{c}{0.94} &
 \multicolumn{2}{c}{0.93} &
 \multicolumn{2}{c}{0.83} & \multicolumn{2}{c}{0.94} & \multicolumn{2}{c|}{0.93}
 &\multicolumn{2}{c}{0.64} &
 \multicolumn{2}{c}{0.68} &
 \multicolumn{2}{c}{0.72} & \multicolumn{2}{c}{0.78} & \multicolumn{2}{c}{0.90} \\
 15 & \multicolumn{2}{c}{0.93} &
 \multicolumn{2}{c}{0.94} &
 \multicolumn{2}{c}{0.88} & \multicolumn{2}{c}{0.95} & \multicolumn{2}{c|}{0.94}
 &\multicolumn{2}{c}{0.80} &
 \multicolumn{2}{c}{0.70} &
 \multicolumn{2}{c}{0.85} & \multicolumn{2}{c}{0.78} & \multicolumn{2}{c}{0.90} \\
 20 & \multicolumn{2}{c}{0.93} &
 \multicolumn{2}{c}{0.93} &
 \multicolumn{2}{c}{0.94} & \multicolumn{2}{c}{0.95} & \multicolumn{2}{c|}{0.93}
 &\multicolumn{2}{c}{0.83} &
 \multicolumn{2}{c}{0.71} &
 \multicolumn{2}{c}{0.85} & \multicolumn{2}{c}{0.81} & \multicolumn{2}{c}{0.90} \\
 25 & \multicolumn{2}{c}{0.93} &
 \multicolumn{2}{c}{0.93} &
 \multicolumn{2}{c}{0.94} & \multicolumn{2}{c}{0.95} & \multicolumn{2}{c|}{0.94}
 &\multicolumn{2}{c}{0.83} &
 \multicolumn{2}{c}{0.77} &
 \multicolumn{2}{c}{0.87} & \multicolumn{2}{c}{0.81} & \multicolumn{2}{c}{0.91} \\
  \bottomrule
 \end{tabular}
 }
 \caption{Comparing F1 scores for feature selection on low feature count sets. f indicates the number of features acquired.}
\label{f1_comp1}
\end{table*}

\begin{figure}[]
    \centering
    \begin{minipage}[t]{0.4\textwidth}
	\centering	
        \includegraphics[width=\linewidth]{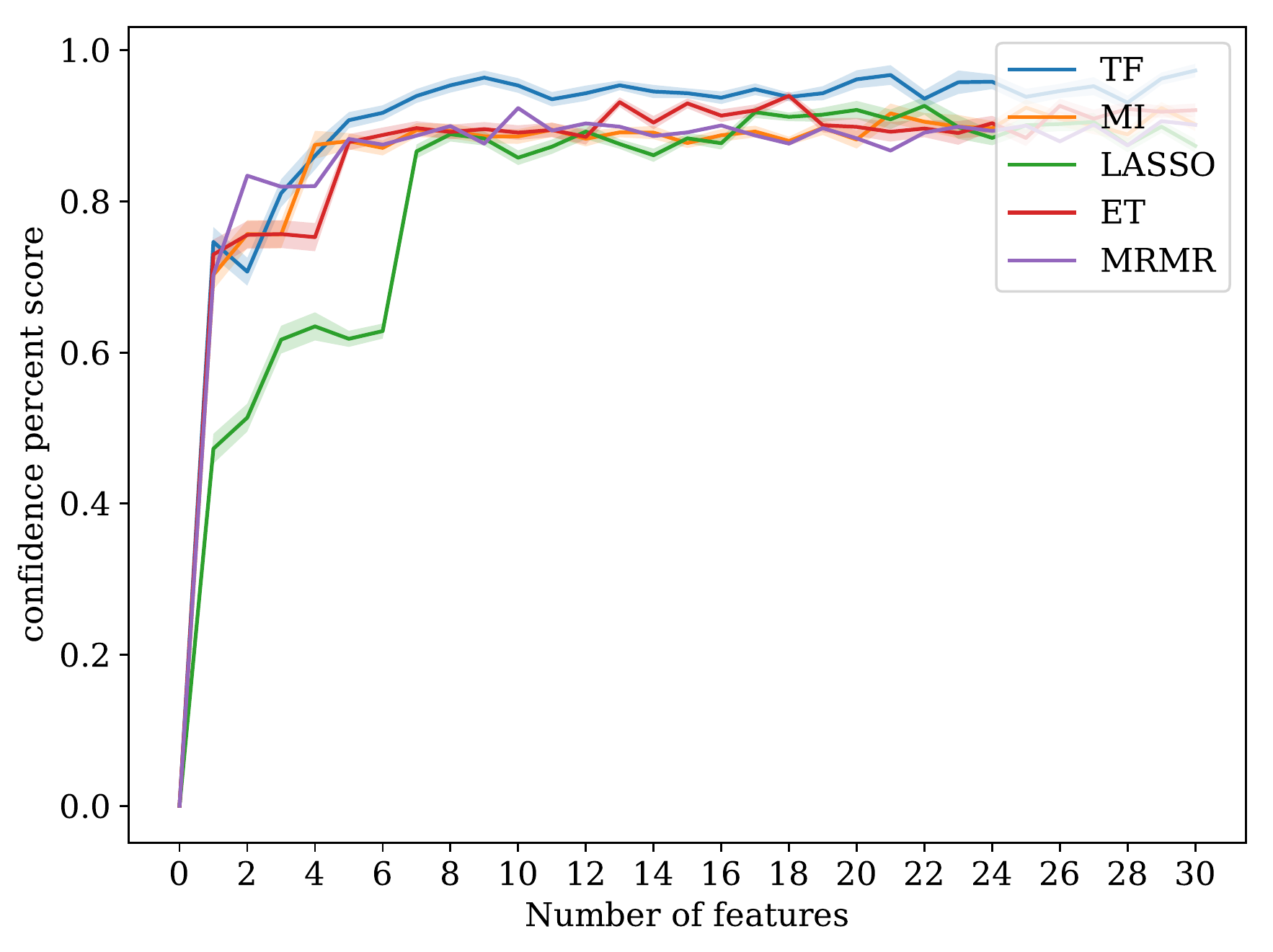}
        \caption{Comparing model confidence in predicting malignant breast cancer. Line thickness indicates variance.}
        \label{fig:breast_cancer_all}
    \end{minipage}
    ~
    \begin{minipage}[t]{0.4\textwidth}
	\centering	
        \includegraphics[width=\linewidth]{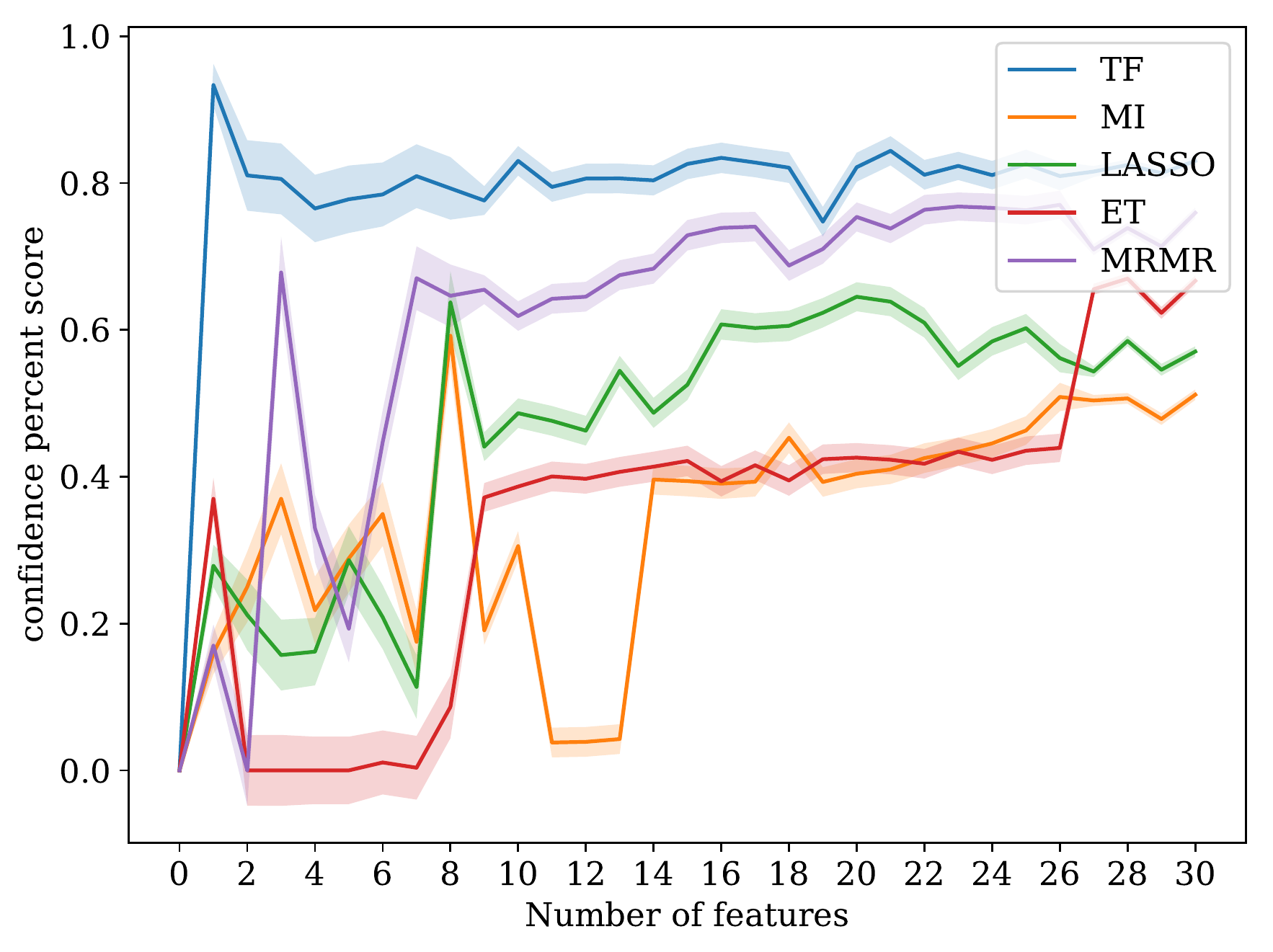}
        \caption{Comparing model confidence in predicting one class out of the Satlog dataset. Line thickness indicates variance.}
        \label{fig:satlog_all}
    \end{minipage}
    \caption*{}
    \vspace{-0.3in}
\end{figure}

\begin{figure}[]
    \vspace{-10pt}
    \centering
    \begin{subfigure}[t]{0.235\textwidth}
	\centering	
        \includegraphics[width=\linewidth]{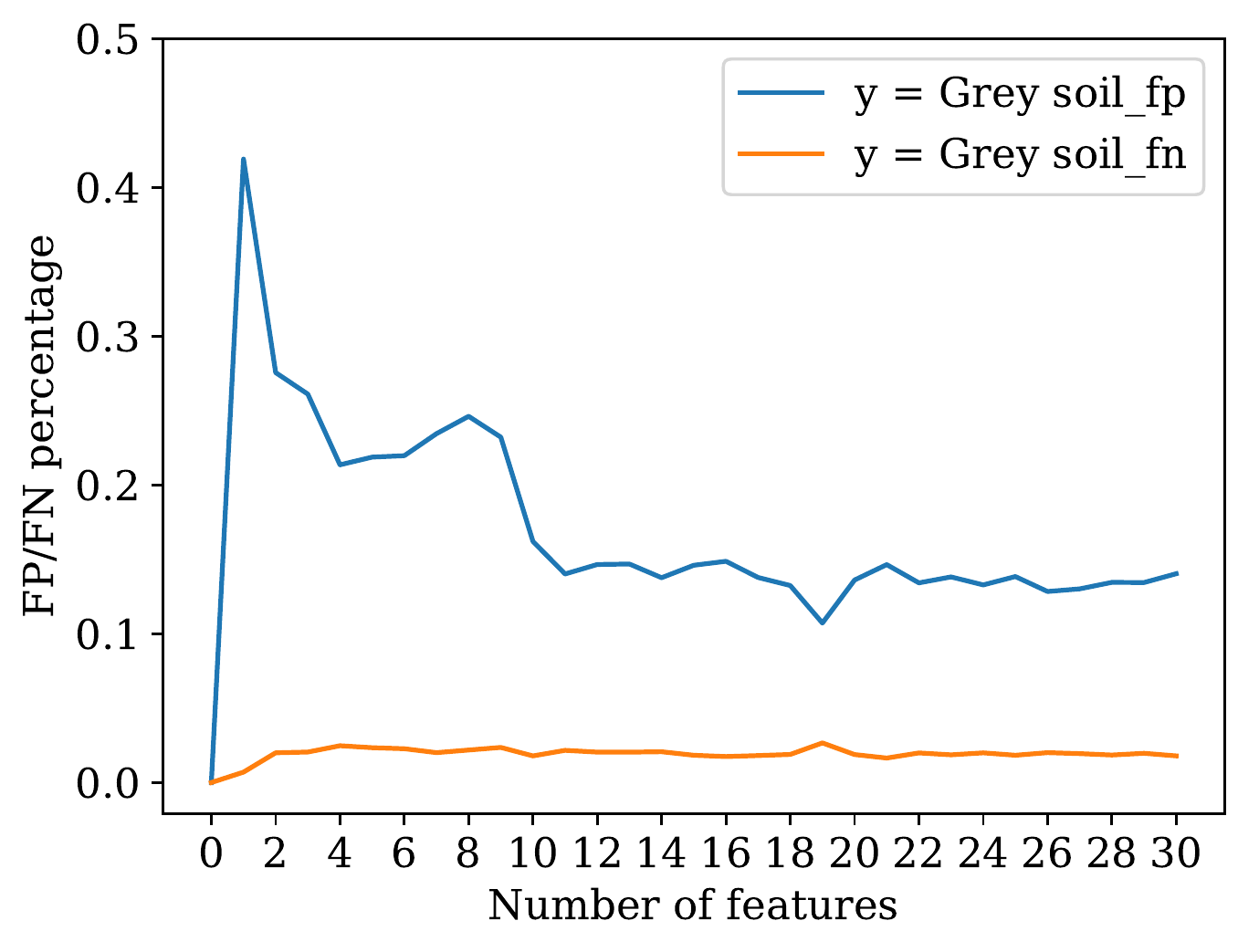}
        \caption{}
        \label{fig:sat_fpfn}
    \end{subfigure}
    \begin{subfigure}[t]{0.235\textwidth}
	\centering	
        \includegraphics[width=\linewidth]{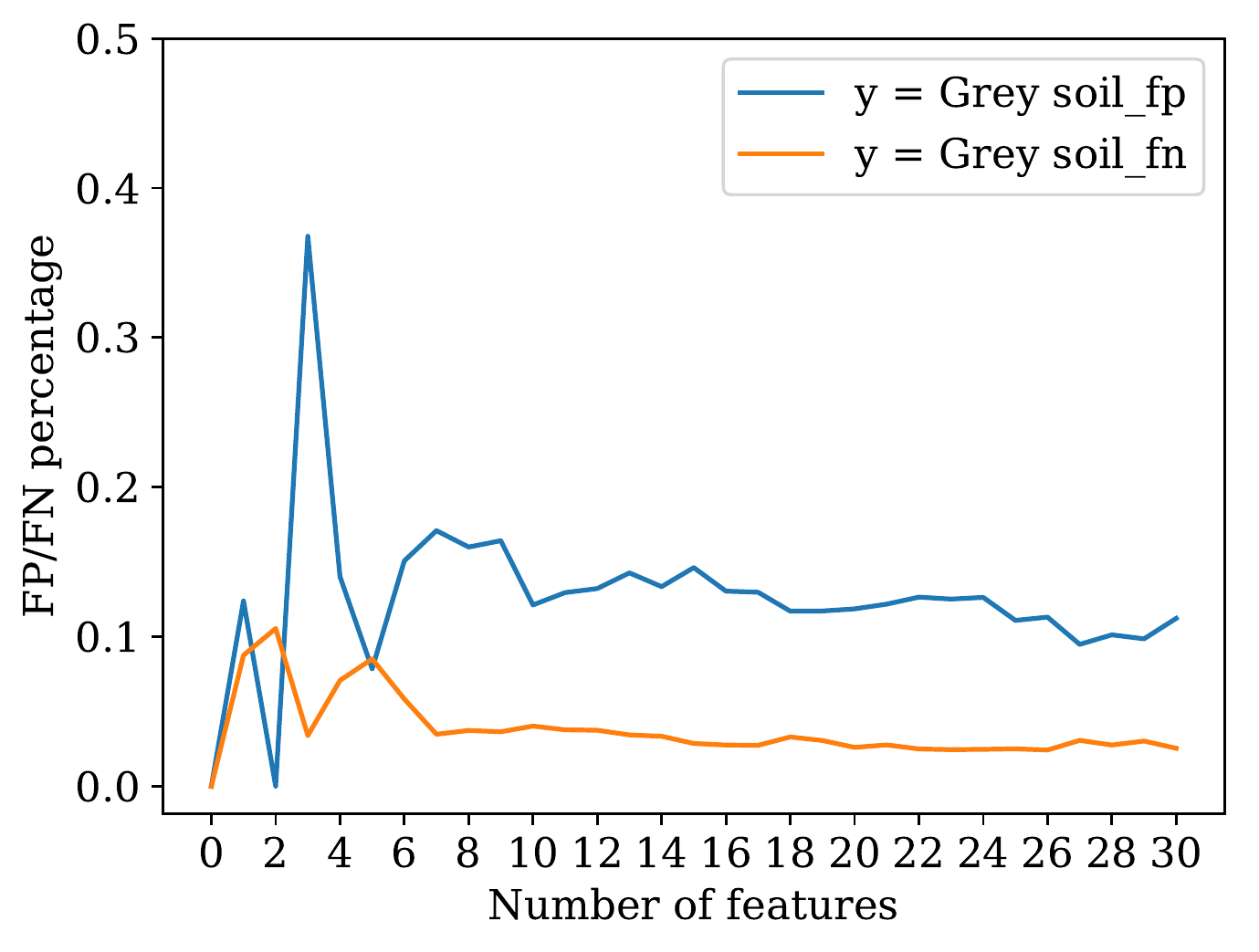}
        \caption{}
        \label{fig:sat_fpfn_mutinfo}
    \end{subfigure}
    \begin{subfigure}[t]{0.235\textwidth}
	\centering	
        \includegraphics[width=\linewidth]{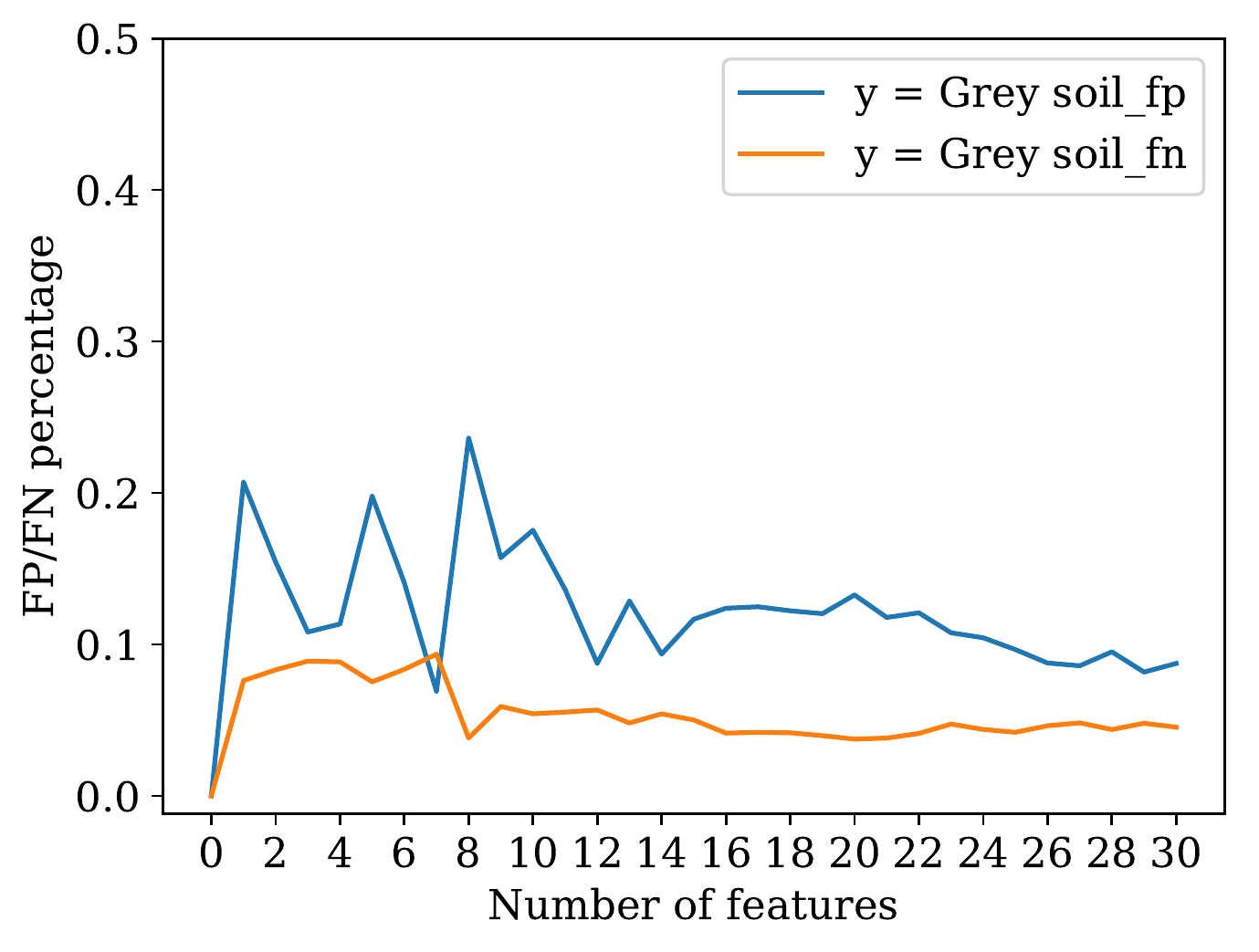}
        \caption{}
        \label{fig:sat_fpfn_lasso}
    \end{subfigure}
    \begin{subfigure}[t]{0.235\textwidth}
	\centering	
        \includegraphics[width=\linewidth]{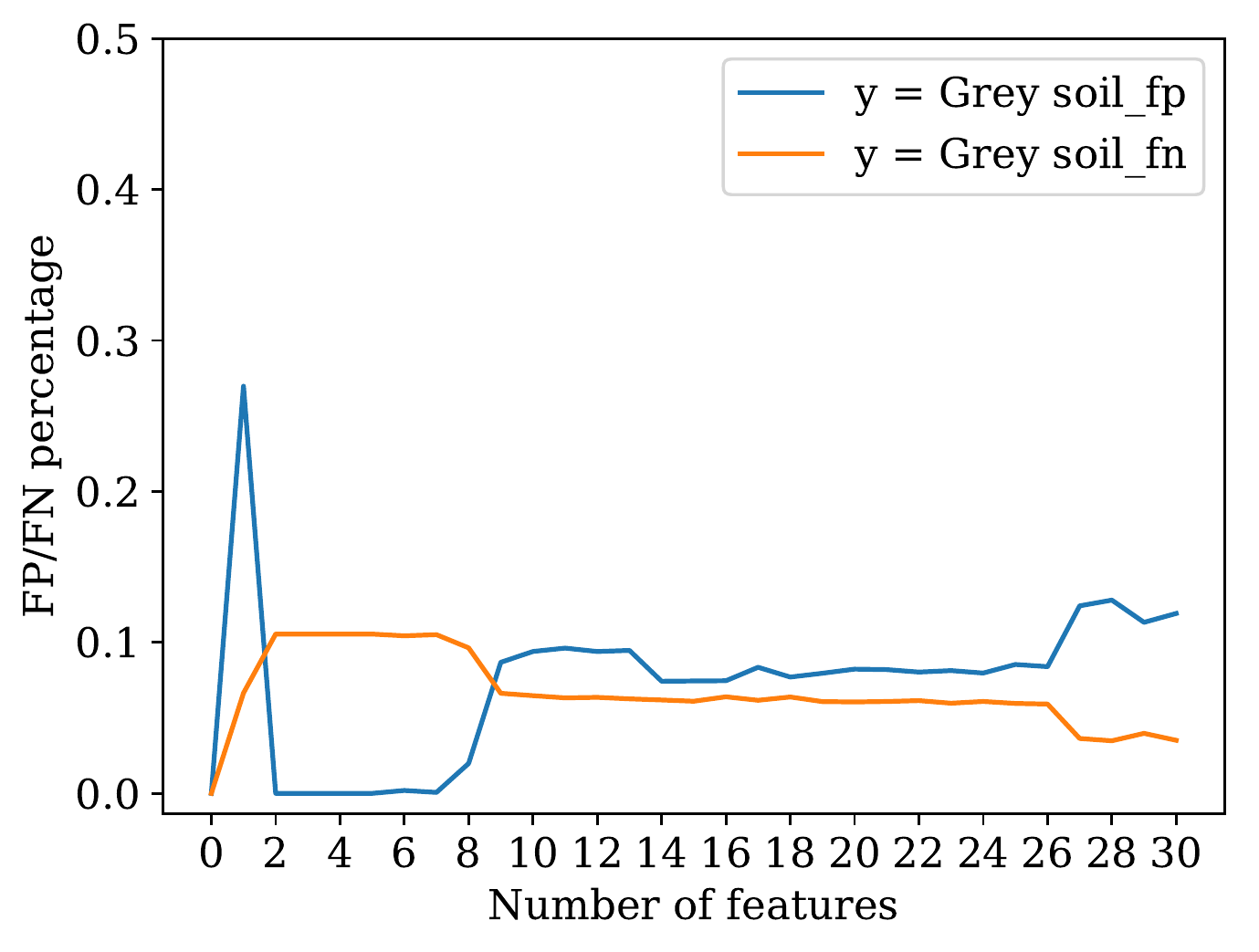}
        \caption{}
        \label{fig:sat_fpfn_extra}
    \end{subfigure}
    \caption{Analysis of UCI Satlog dataset comparing FP/FN rates as features are acquired. \ref{fig:sat_fpfn} shows evaluation using our approach, \ref{fig:sat_fpfn_mutinfo} shows evaluation using mRMR "MIQ" method, \ref{fig:sat_fpfn_lasso} shows evaluation using Lasso method, \ref{fig:sat_fpfn_extra} shows evaluation using Extra trees}
    \label{fig:sat_fpfn_all}
    \vspace{-0.2in}
\end{figure}

On data sets with low feature count and little missingness, our method was able to achieve better overall scores, faster than the comparable methods For a specific target value. As can be seen in Table \ref{f1_comp1}.

Figure \ref{fig:breast_cancer_all} shows confidence trend for acquiring 30 features on the Breast Cancer Wisconsin dataset using our method, comparing specific target confidence in the 4 compared models. In this case, our model can be seen on par with the confidence achieved by the Mutual information and Extra trees methods. As can be seen in Table \ref{f1_comp1}, our model is able to produce slightly better F1 scores, indicating a faster false positive and false negative reduction. The breast cancer dataset proved to be a relatively simple prediction problem, it can be seen that all methods performed relatively well, achieving good model confidence and F1 scores. 

Figure \ref{fig:satlog_all} shows confidence trend for acquiring 30 features on the Satlog dataset using our and compared methods. Here we can see our target-focused method is able to achieve a more confident model faster, as well as a better overall F1 score. In this case, the target of focus chosen appeared to be the hardest target to model out of the available targets, since all compared models struggled to find features that best model the data, in addition to it being one of the minority classes. Despite that, our model has gained the most confidence, while using a low number of features. 
Figures \ref{fig:sat_fpfn} to \ref{fig:sat_fpfn_extra} show the FP/FN evaluation over the Satlog dataset. We see our method shows a consistent non-volatile decline in FP rates while maintaining a low FN rate throughout.

On Table \ref{f1_comp2} we can see the F1 scores of the different feature selection methods compared on high feature count datasets, with missing values and lots of features. Here our model heuristic is evaluated on publicly available real-world healthcare data. Our method, being specific target aware, is able to pick out a good subset of the features consistently, consequently using fewer features that in turn contribute most to maximizing the selected target class in focus.

\begin{figure}[H]
    \centering
    \begin{minipage}[]{0.4\textwidth}
	\centering	
        \includegraphics[width=\linewidth]{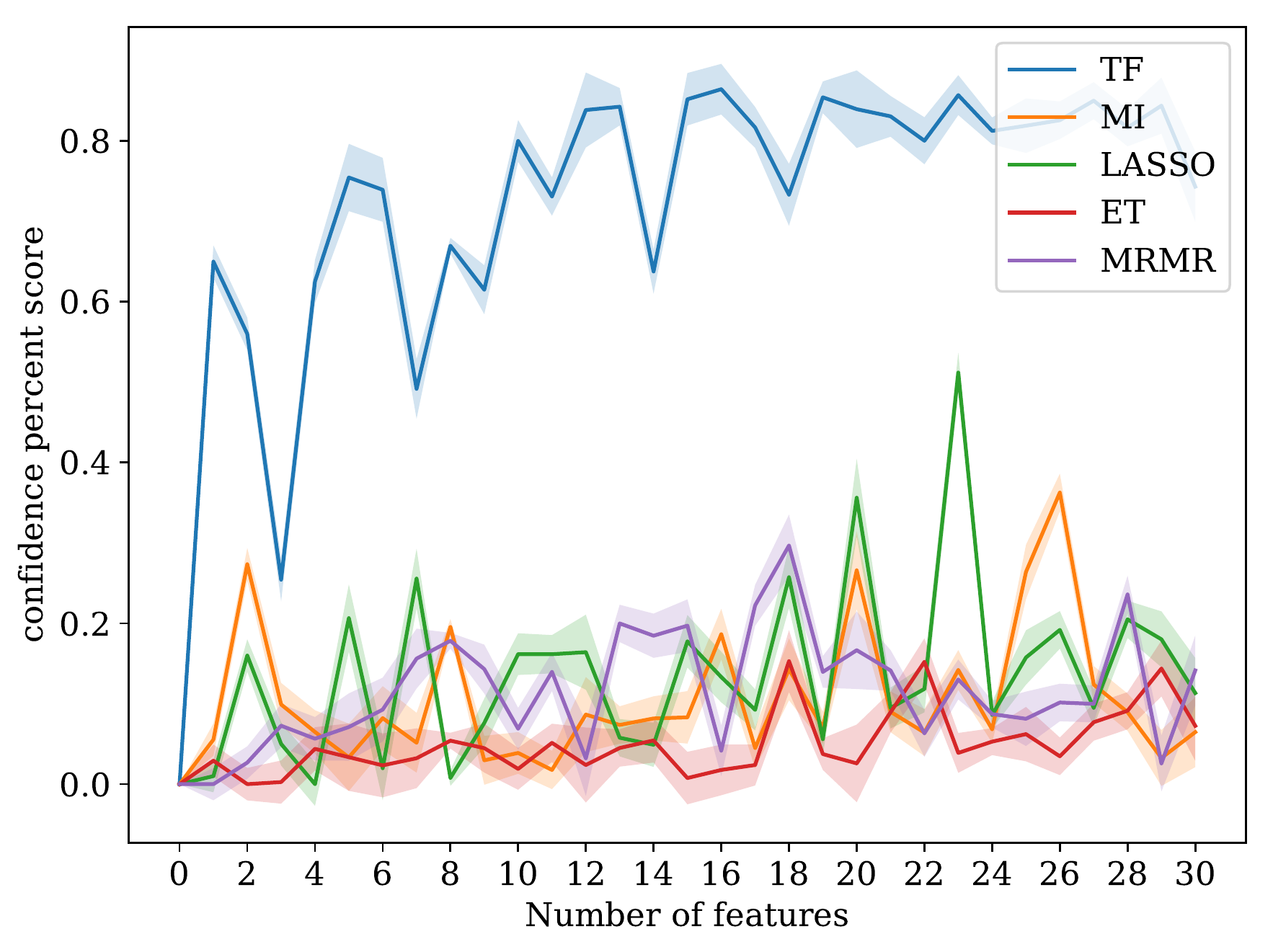}
        \caption{Comparing model confidence in predicting congestive heart failure. Line thickness indicates variance.}
        \label{fig:heart_all}
    \end{minipage}
    ~
    \begin{minipage}[]{0.4\textwidth}
	\centering	
        \includegraphics[width=\linewidth]{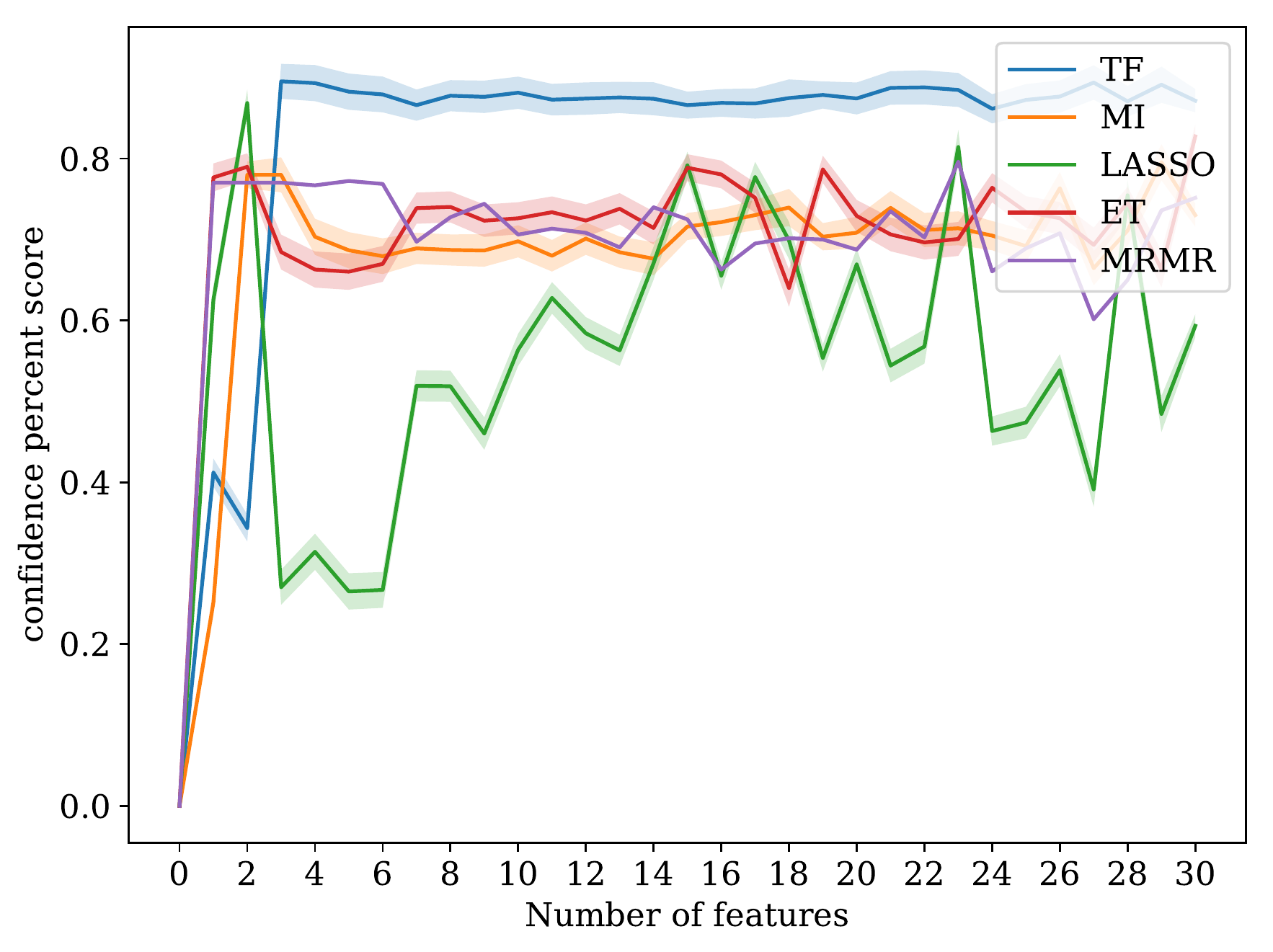}
        \caption{Comparing model confidence in predicting diabetes. Line thickness indicates variance.}
        \label{fig:diabetes_all}
    \end{minipage}
    \caption*{}
\end{figure}

\begin{figure}[H]
    \vspace{-35pt}
    \centering
    \begin{subfigure}[]{0.235\textwidth}
	\centering	
        \includegraphics[width=\linewidth]{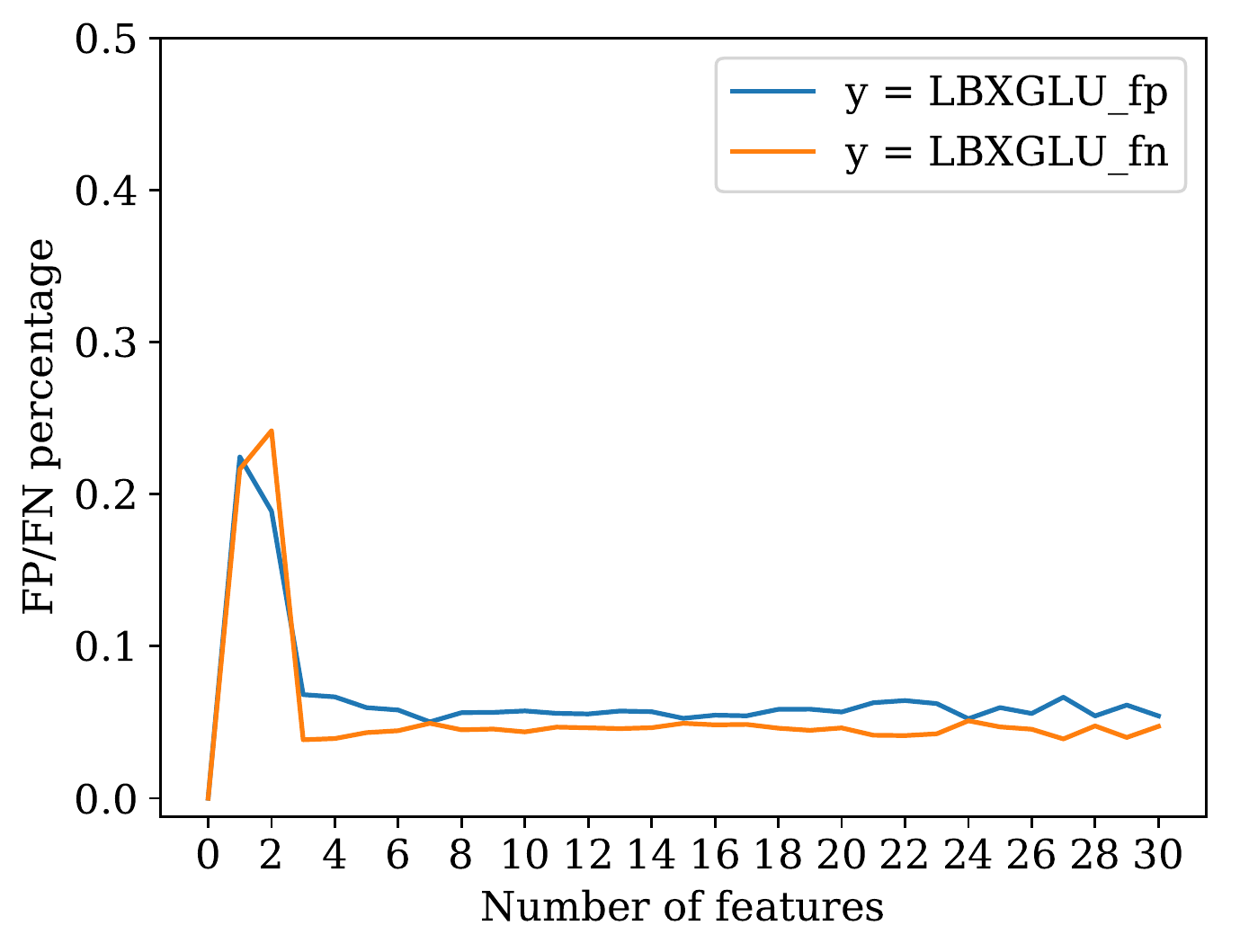}
        \caption{}
        \label{fig:diabetes_fpfn}
    \end{subfigure}
    \begin{subfigure}[]{0.235\textwidth}
	\centering	
        \includegraphics[width=\linewidth]{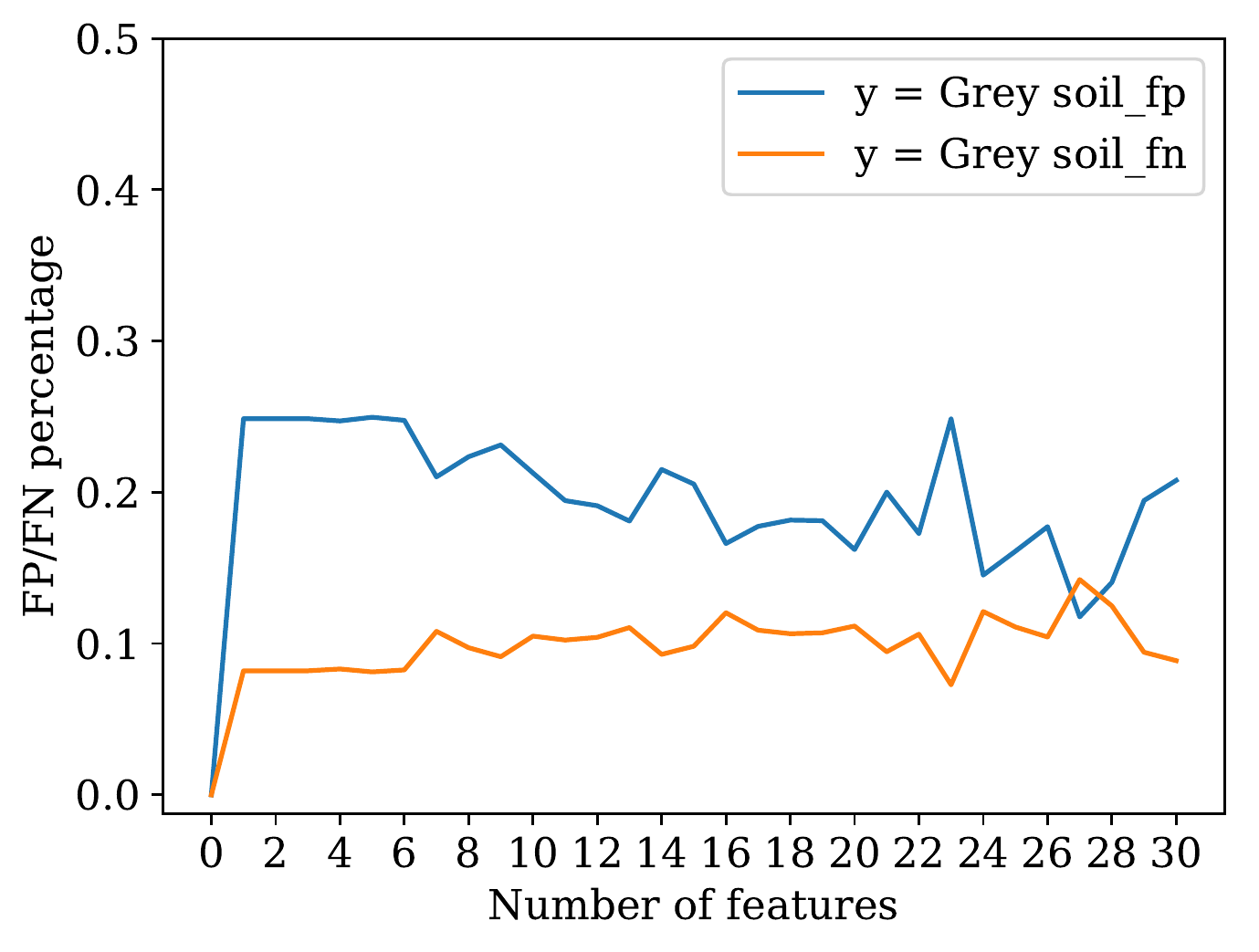}
        \caption{}
        \label{fig:diabetes_fpfn_mutinfo}
    \end{subfigure}
    \begin{subfigure}[]{0.235\textwidth}
	\centering	
        \includegraphics[width=\linewidth]{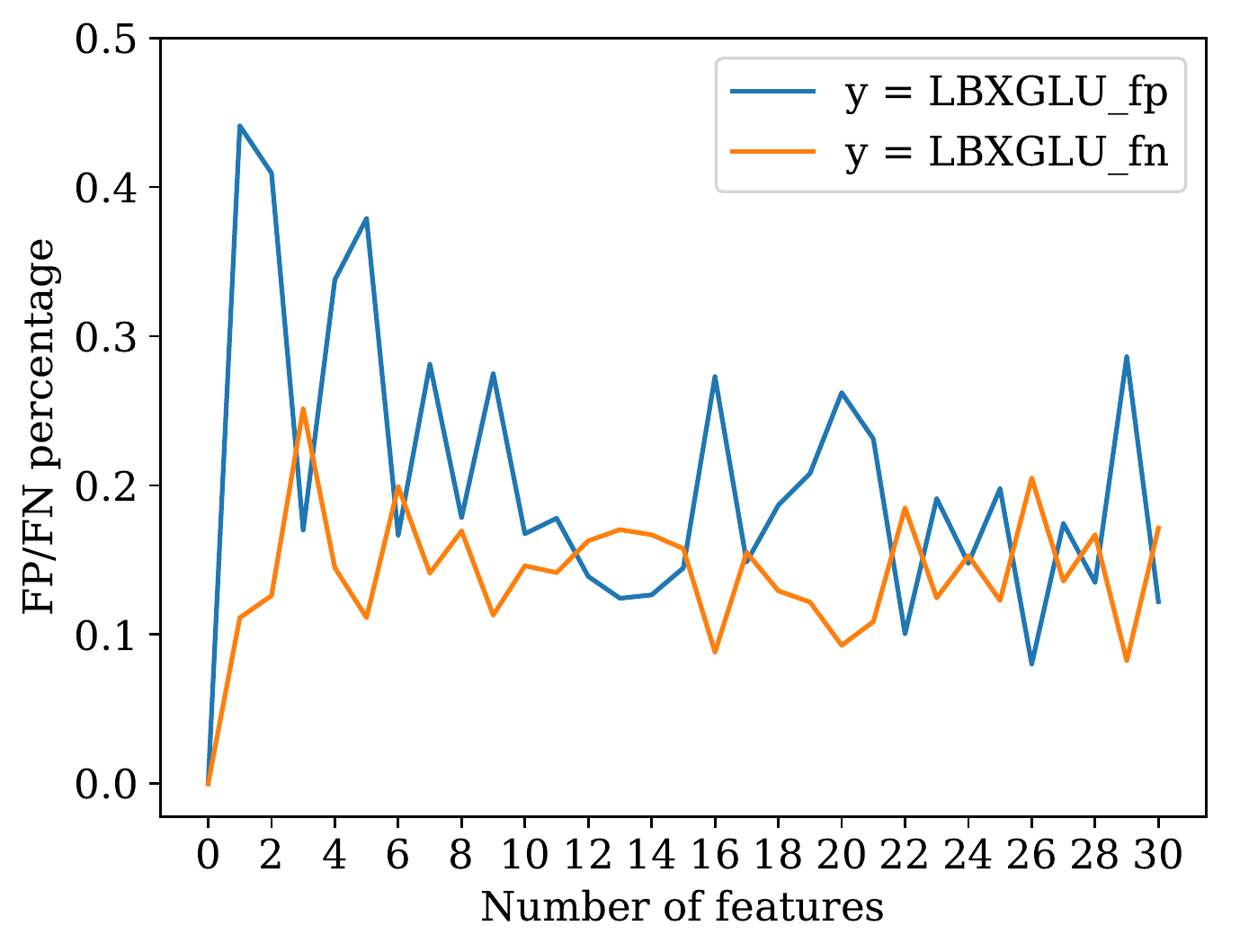}
        \caption{}
        \label{fig:diabetes_fpfn_lasso}
    \end{subfigure}
    \begin{subfigure}[]{0.235\textwidth}
	\centering	
        \includegraphics[width=\linewidth]{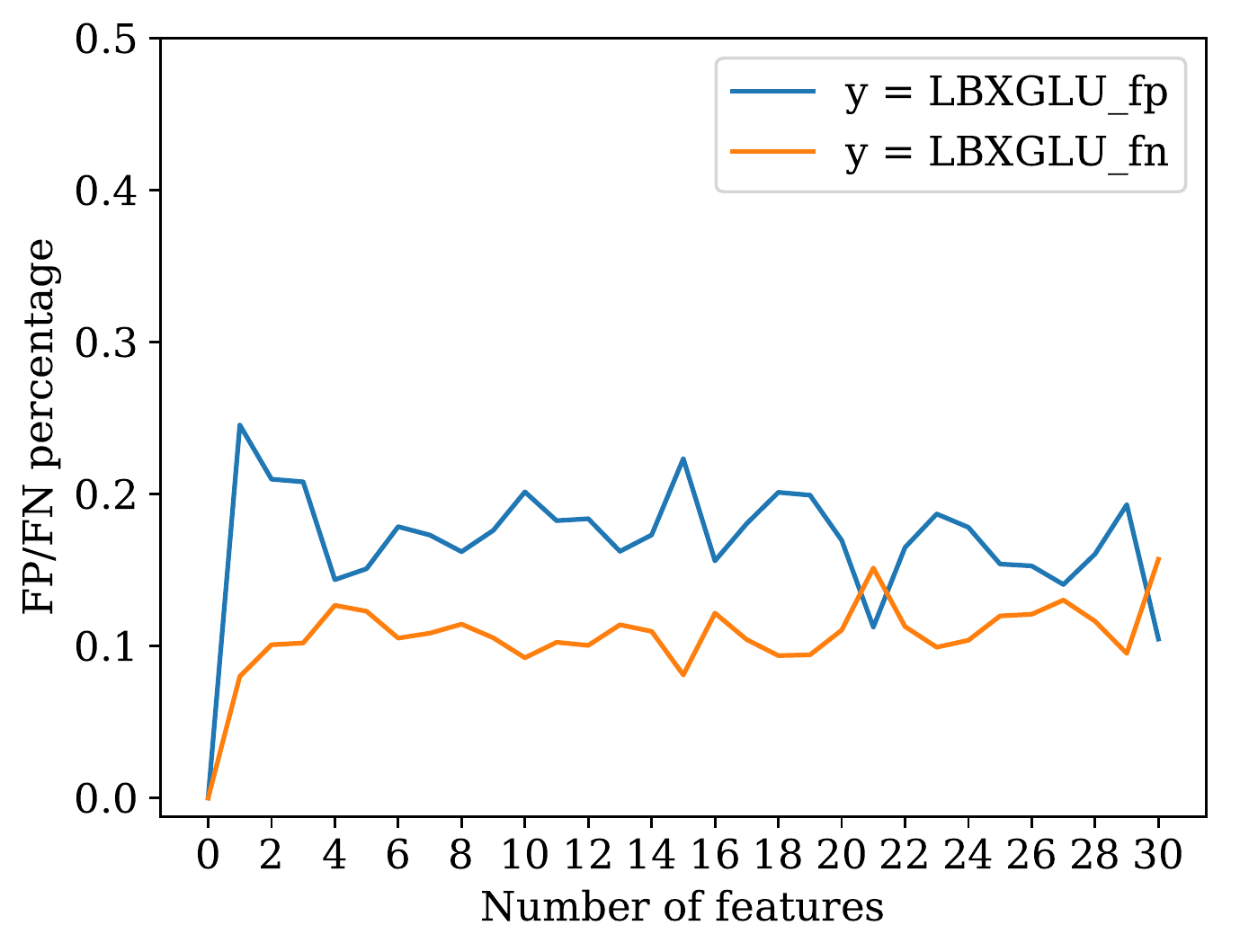}
        \caption{}
        \label{fig:diabetes_fpfn_extra}
    \end{subfigure}
    \caption{Analysis of NHANES Diabetes constructed dataset comparing FP/FN rates as features are acquired. \ref{fig:diabetes_fpfn} shows evaluation using our approach, \ref{fig:diabetes_fpfn_mutinfo} shows evaluation using mRMR "MIQ" method, \ref{fig:diabetes_fpfn_lasso} shows evaluation using Lasso method, \ref{fig:diabetes_fpfn_extra} shows evaluation using Extra trees.}
    \label{fig:diabetes_fpfn_all}
    \vspace{-0.1in}
\end{figure}

As can be seen in the diabetes confidence evaluation in Figure \ref{fig:diabetes_all}, and in the FP/FN evaluation in Figures \ref{fig:diabetes_fpfn} to \ref{fig:diabetes_fpfn_extra}, our method outperformed the compared methods in minimizing FP and FN scores quickly, in addition to achieving a consistent amount of confidence in the target of interest relatively fast. It can be seen in Figure \ref{fig:diabetes_all} that all comparable methods achieve a high amount of confidence quicker than our target-focused method. However, comparing Figures \ref{fig:diabetes_fpfn} to \ref{fig:diabetes_fpfn_extra} we can see our model minimizes false positive and false negative scores quicker and therefore receives higher F1 scores. 
 
Confidence evaluation for the heart disease dataset can be seen in Figure \ref{fig:heart_all}. Our model is gaining confidence using fewer features as before and keeps a relatively increasing trend of confidence. Other models failed to increase their confidence significantly as this was the hardest task, with multiple targets and high dimensionality. 

\begin{table*}[t]
\centering
\resizebox{0.85\textwidth}{!}{
 \begin{tabular}{ c | c c c c c c c c c c c c c c c c c c c c c c c c c c c c} 
 \toprule
  \multicolumn{21}{c}{\textbf{F1 scores}} \\
 \hline
 & \multicolumn{10}{c|}{\textbf{NHANES Diabetes}} & \multicolumn{10}{c}{\textbf{NHANES Heart}} \\
  \hline
 \textbf{f} &\multicolumn{2}{c}{\textbf{MI}} &
 \multicolumn{2}{c}{\textbf{mRMR}} &
 \multicolumn{2}{c}{\textbf{Lasso}} & \multicolumn{2}{c}{\textbf{Extra Trees}} & \multicolumn{2}{c|}{\textbf{TF}}
 &\multicolumn{2}{c}{\textbf{MI}} &
 \multicolumn{2}{c}{\textbf{mRMR}} &
 \multicolumn{2}{c}{\textbf{Lasso}} & \multicolumn{2}{c}{\textbf{Extra Trees}} & \multicolumn{2}{c}{\textbf{TF}} \\
  \hline
 5 & \multicolumn{2}{c}{0.76} &
 \multicolumn{2}{c}{0.79} &
 \multicolumn{2}{c}{0.61} & \multicolumn{2}{c}{0.79} & \multicolumn{2}{c|}{0.92}
 &\multicolumn{2}{c}{0.66} &
 \multicolumn{2}{c}{0.55} &
 \multicolumn{2}{c}{0.28} & \multicolumn{2}{c}{0.39} & \multicolumn{2}{c}{0.72} \\
 10 & \multicolumn{2}{c}{0.78} &
 \multicolumn{2}{c}{0.77} &
 \multicolumn{2}{c}{0.73} & \multicolumn{2}{c}{0.79} & \multicolumn{2}{c|}{0.92}
 &\multicolumn{2}{c}{0.31} &
 \multicolumn{2}{c}{0.48} &
 \multicolumn{2}{c}{0.59} & \multicolumn{2}{c}{0.57} & \multicolumn{2}{c}{0.78} \\
 15 & \multicolumn{2}{c}{0.80} &
 \multicolumn{2}{c}{0.79} &
 \multicolumn{2}{c}{0.77} & \multicolumn{2}{c}{0.77} & \multicolumn{2}{c|}{0.92}
 &\multicolumn{2}{c}{0.78} &
 \multicolumn{2}{c}{0.69} &
 \multicolumn{2}{c}{0.56} & \multicolumn{2}{c}{0.32} & \multicolumn{2}{c}{0.87} \\
 20 & \multicolumn{2}{c}{0.78} &
 \multicolumn{2}{c}{0.80} &
 \multicolumn{2}{c}{0.76} & \multicolumn{2}{c}{0.74} & \multicolumn{2}{c|}{0.92}
 &\multicolumn{2}{c}{0.68} &
 \multicolumn{2}{c}{0.63} &
 \multicolumn{2}{c}{0.61} &
 \multicolumn{2}{c}{0.68} & \multicolumn{2}{c}{0.87} \\
 25 & \multicolumn{2}{c}{0.76} &
 \multicolumn{2}{c}{0.80} &
 \multicolumn{2}{c}{0.64} & \multicolumn{2}{c}{0.77} & \multicolumn{2}{c|}{0.92}
 &\multicolumn{2}{c}{0.85} &
 \multicolumn{2}{c}{0.76} &
 \multicolumn{2}{c}{0.66} &
 \multicolumn{2}{c}{0.47} & \multicolumn{2}{c}{0.86} \\
  \bottomrule
 \end{tabular}
 }
 \caption{Comparing F1 scores for feature selection on high feature count sets. f indicates the number of features acquired.}
\label{f1_comp2}
\end{table*}

Since the other feature selection models are unaware of the single target uncertainty in the model, the results obtained by the compared models could be largely dependent on the distribution of targets. I.e: the selected target of focus might get better results if it is also the majority target. All models compared were able to find features to construct an efficient frugal model on at least one of the sets, but our method has shown higher consistency across all sets. While real-world health data is normally sparse and feature-rich, we can see that even on smaller datasets with fewer features, our method provides a good heuristic as to the value of features when acquired towards a single target.

\section{Conclusion}
In this paper, we have investigated the approach of acquiring features based on a specific target of interest out of two or more targets. We see a frugal approach as an important addition to the process of feature selection, especially as data availability grows dramatically, and utilization of data remains somewhat inefficient, particularly in the domain of healthcare. 
We have discussed the application of our target-focused approach to both well-known sources of machine learning datasets, as well as real-world public healthcare data converted into datasets. On these, we have clearly demonstrated the value of having a target-aware method to feature selection, as compared to feature selection methods that are target-agnostic. We have introduced a Bayesian confidence based scoring mechanism, that we proceeded to show is robust in both scalability and consistency on different types of datasets. Practically, we were able to minimize uncertainty in a specific target of interest with a minimal budget, while minimizing the general uncertainty, false positive, and false negative rates.

%


\bibliographystyle{plain}
\bibliography{neurips_bayesian}

\end{document}


\maketitle
\appendix
\section{Additional Comparisons}
Additional comparison provided for target focused model confidence plotted with the general confidence scores. Line thickness denotes variance as in the paper.

\subsubsection{Breast Cancer}
\begin{figure}[H]
    \centering
    \begin{minipage}[t]{0.3\textwidth}
	\centering	
        \includegraphics[width=\linewidth]{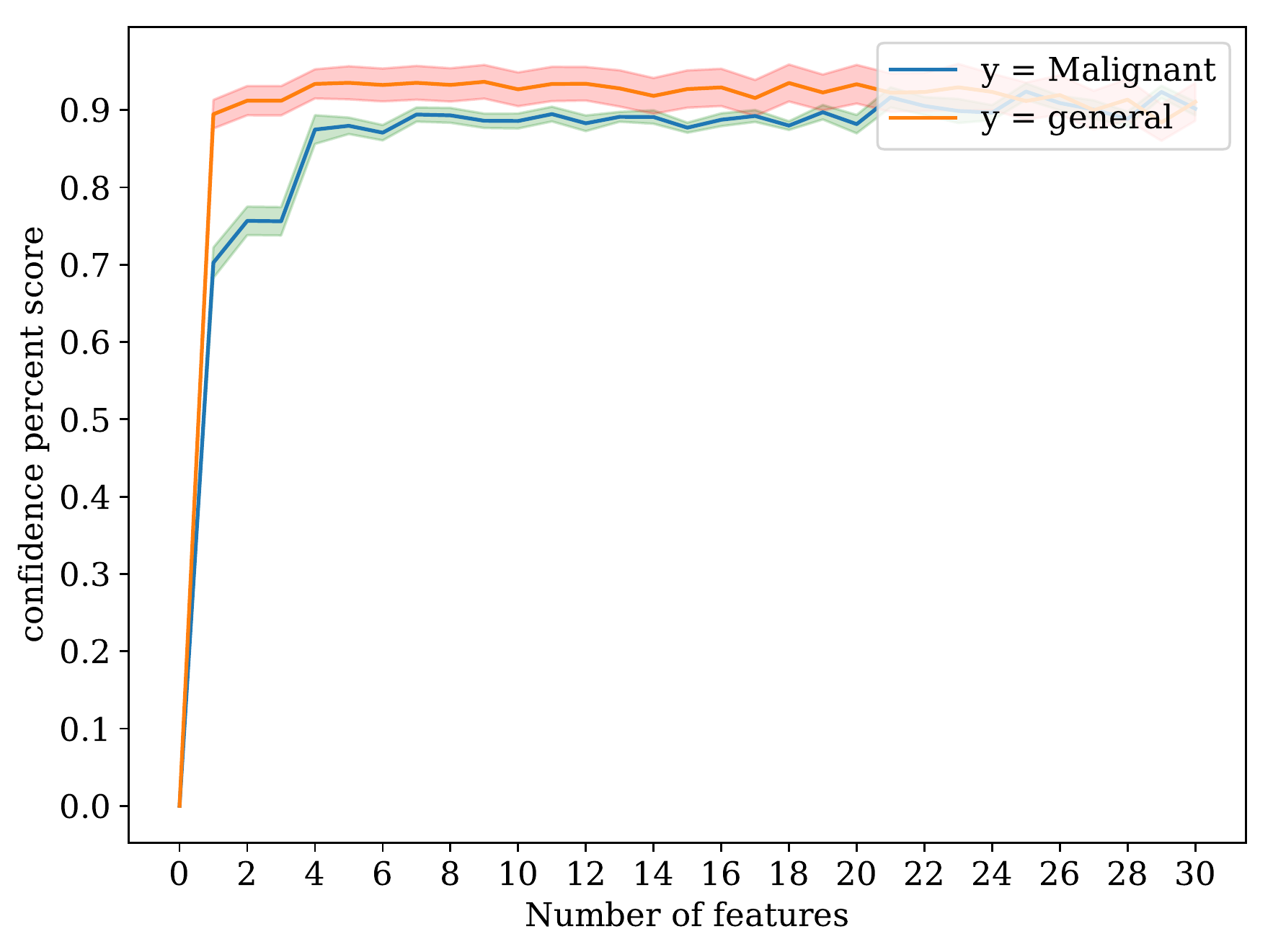}
        \label{fig:breast_cancer_all}
        \caption{Non-target-specific confidence in for mutual information on the breast cancer dataset}
    \end{minipage}
    ~
    \begin{minipage}[t]{0.3\textwidth}
	\centering	
        \includegraphics[width=\linewidth]{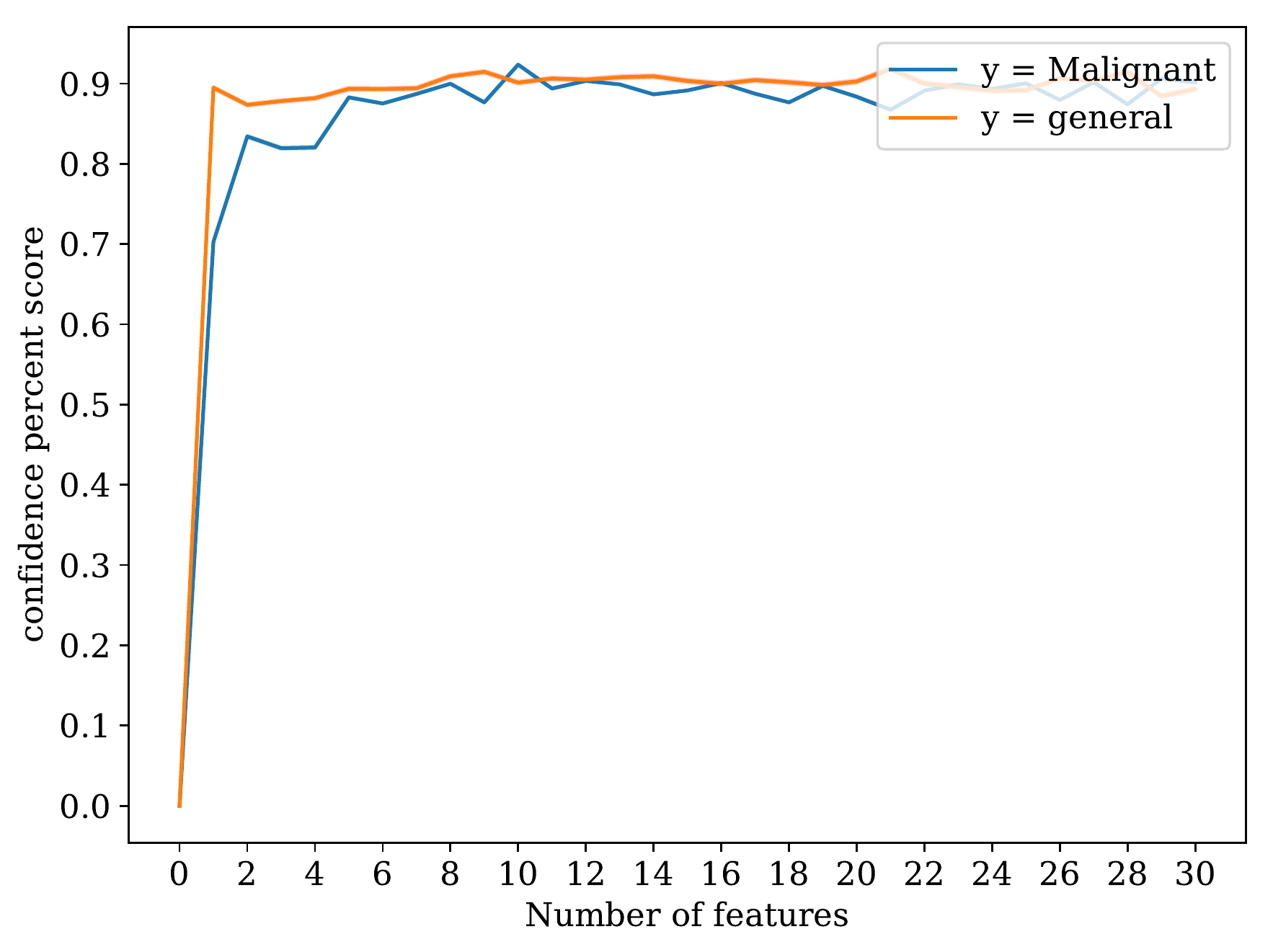}
        \label{fig:satlog_all}
        \caption{Non-target-specific confidence in for mRMR on the breast cancer dataset}
    \end{minipage}
    ~
    \begin{minipage}[t]{0.3\textwidth}
	\centering	
        \includegraphics[width=\linewidth]{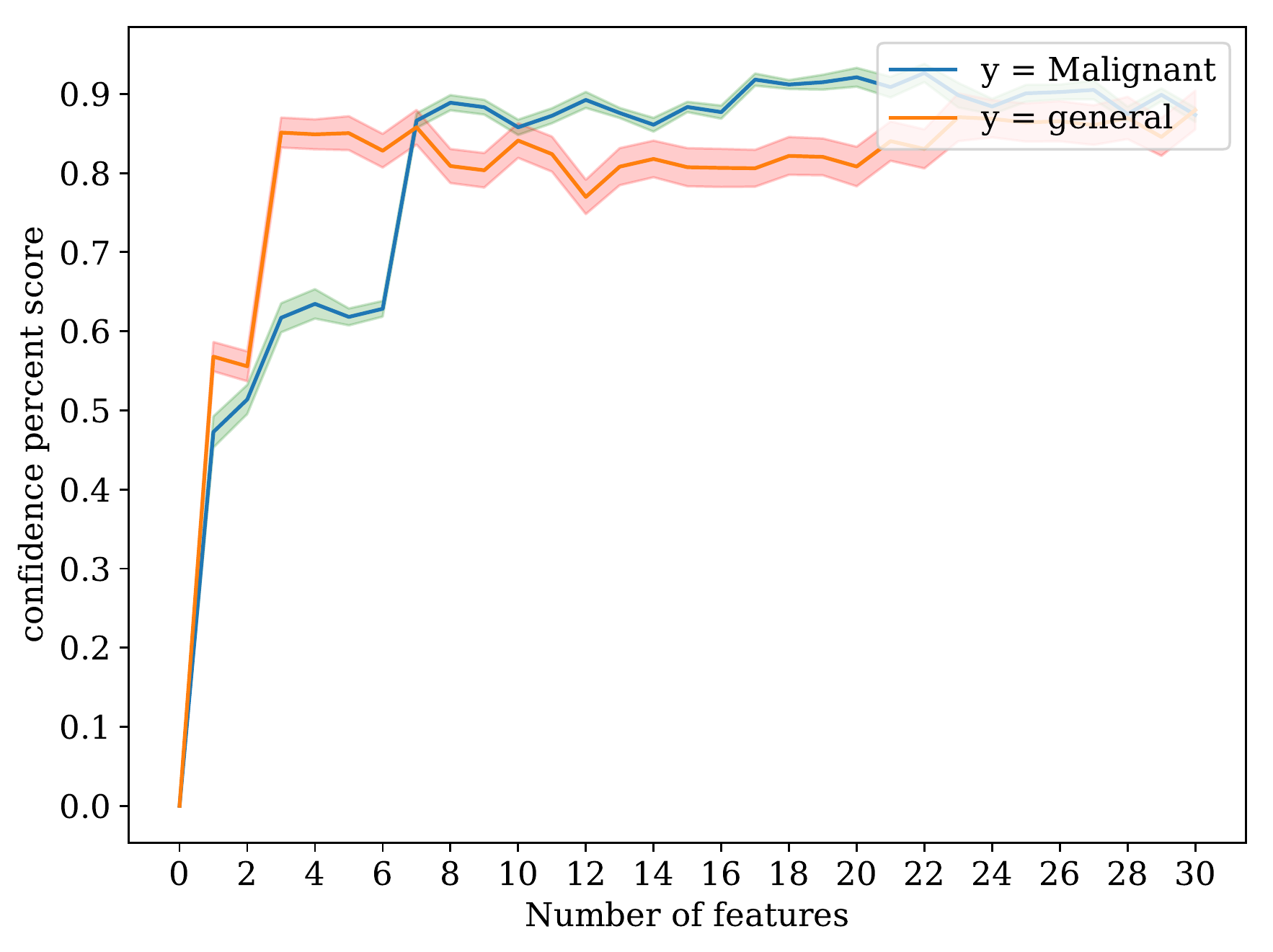}
        \label{fig:breast_cancer_all}
        \caption{Non-target-specific confidence in for Lasso on the breast cancer dataset}
    \end{minipage}
    ~
    \begin{minipage}[t]{0.3\textwidth}
	\centering	
        \includegraphics[width=\linewidth]{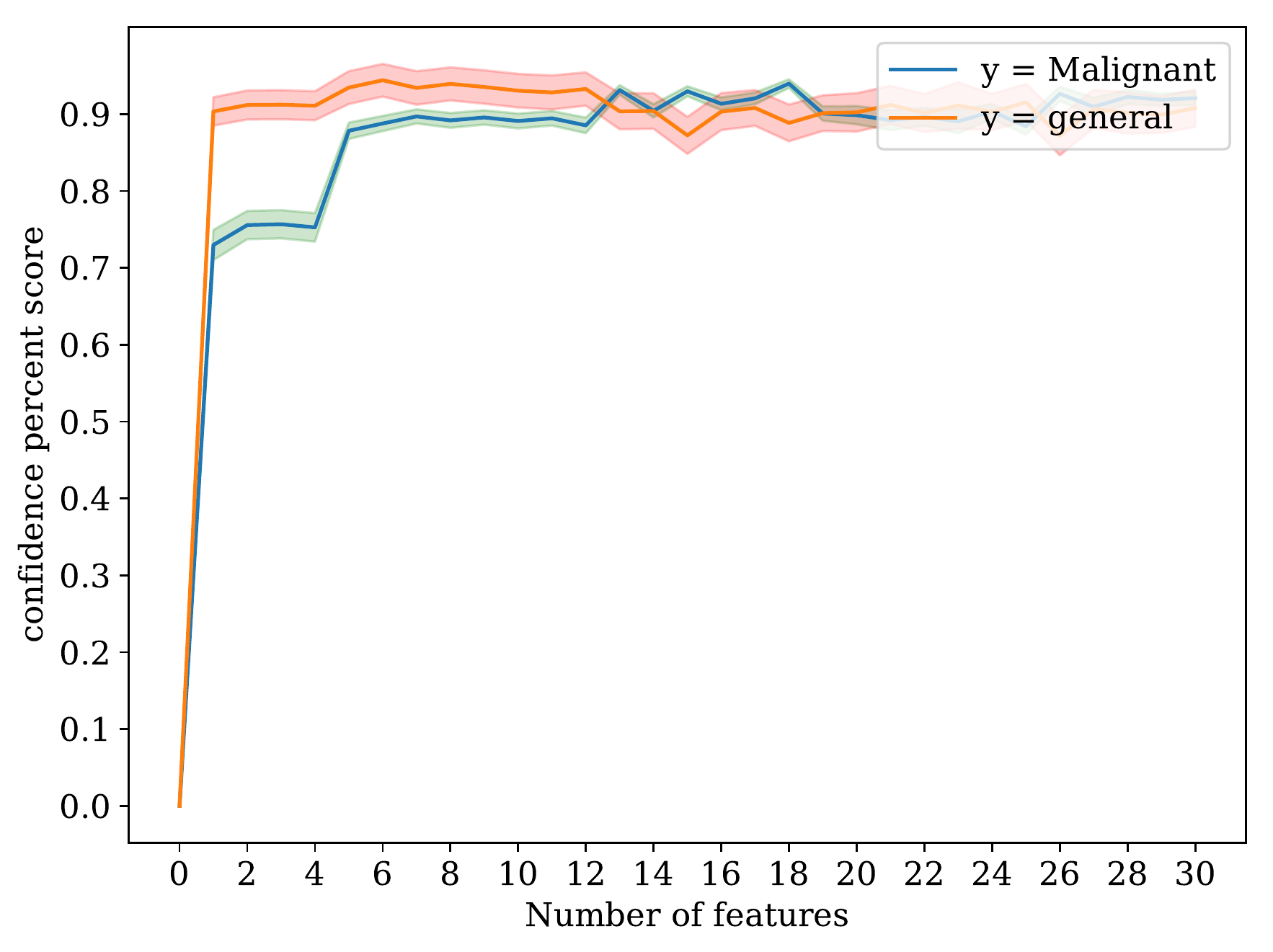}
        \label{fig:satlog_all}
        \caption{Non-target-specific confidence in for Extra Trees on the breast cancer dataset}
    \end{minipage}
    ~
    \begin{minipage}[t]{0.3\textwidth}
	\centering	
        \includegraphics[width=\linewidth]{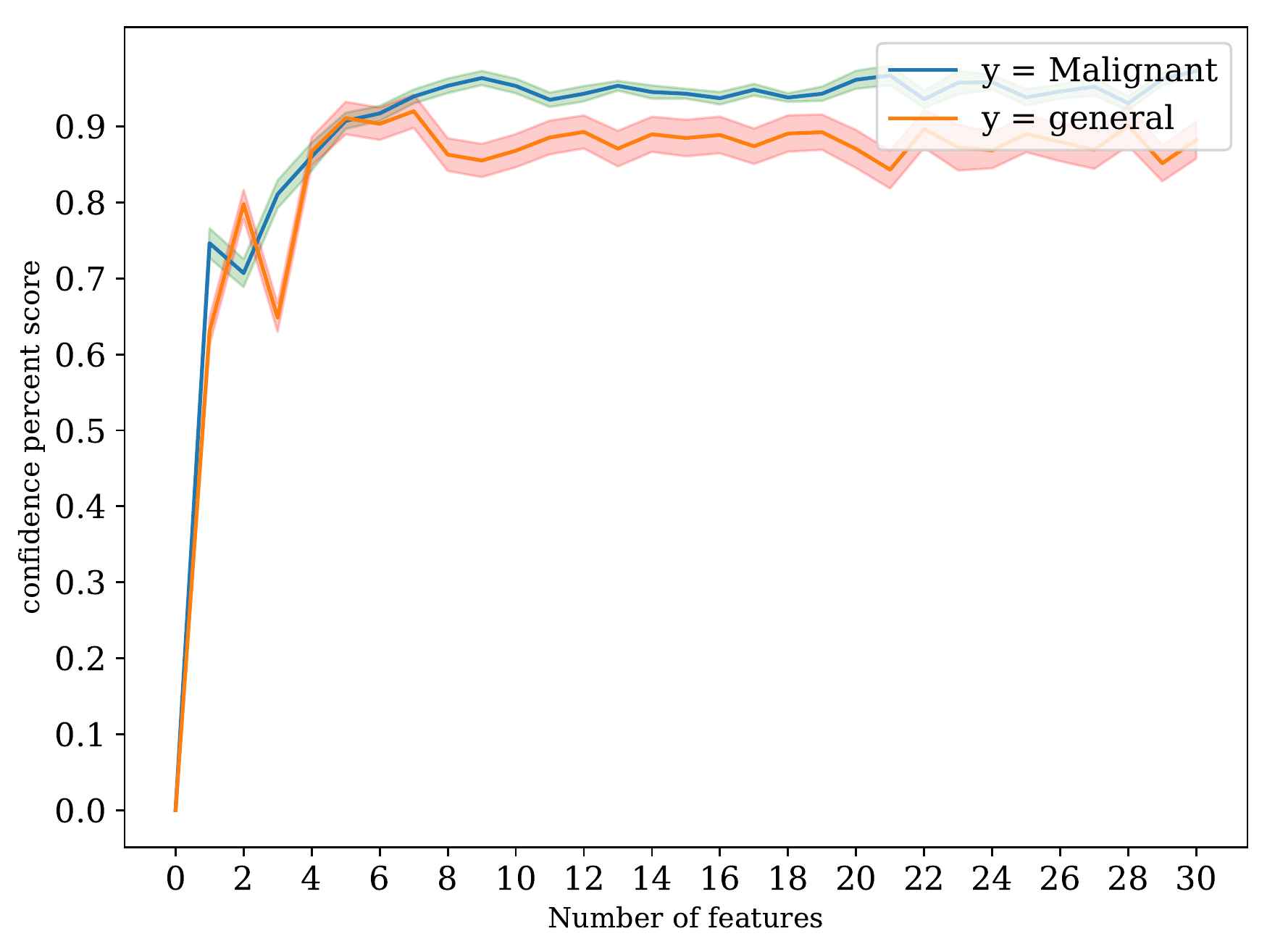}
        \label{fig:satlog_all}
        \caption{Non-target-specific confidence in for Target Focused on the breast cancer dataset}
    \end{minipage}
\end{figure}

\subsubsection{Satlog}
\begin{figure}[H]
    \centering
    \begin{minipage}[t]{0.3\textwidth}
	\centering	
        \includegraphics[width=\linewidth]{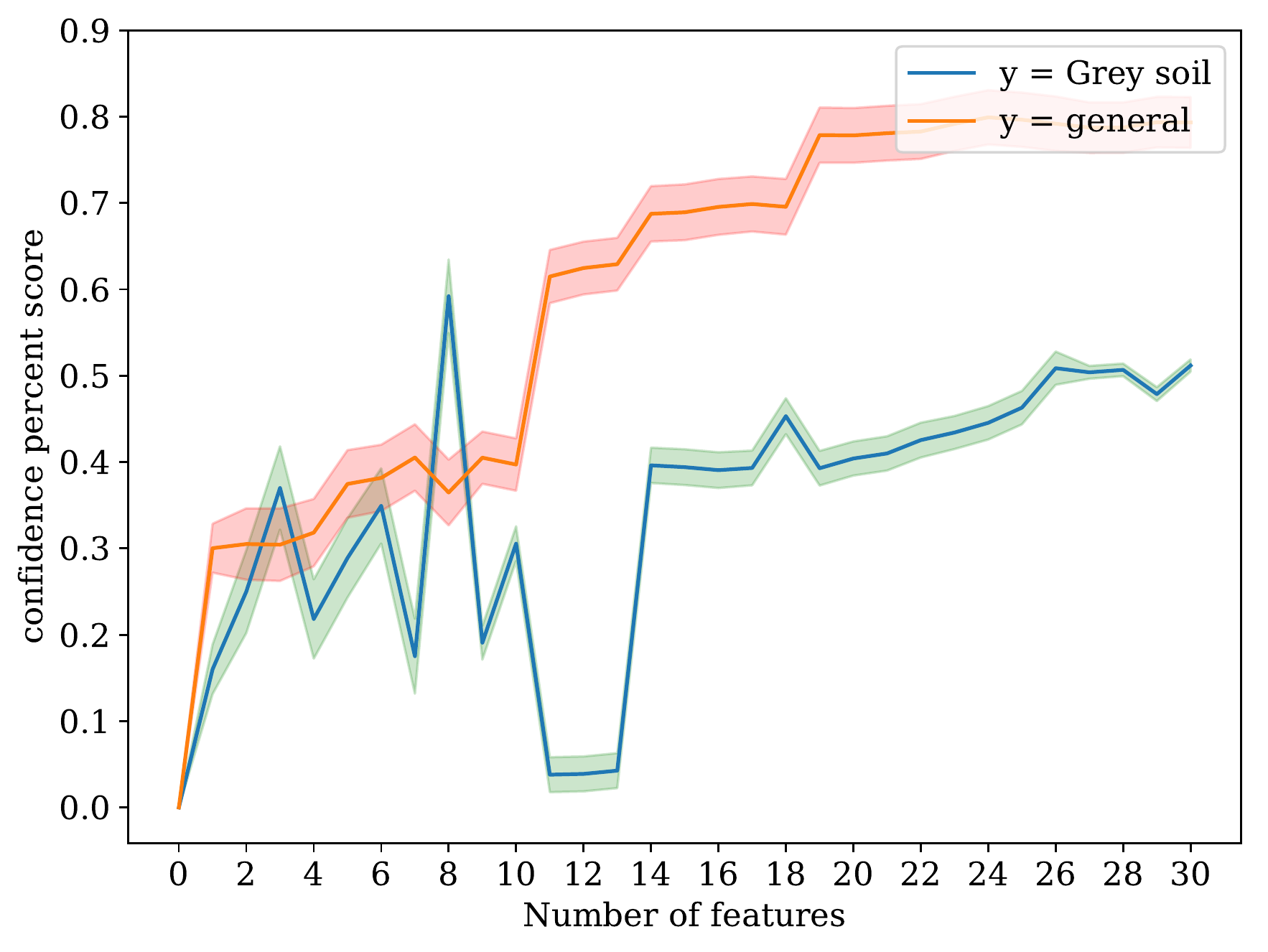}
        \label{fig:breast_cancer_all}
        \caption{Non-target-specific confidence in for mutual information on the Satlog dataset}
    \end{minipage}
    ~
    \begin{minipage}[t]{0.3\textwidth}
	\centering	
        \includegraphics[width=\linewidth]{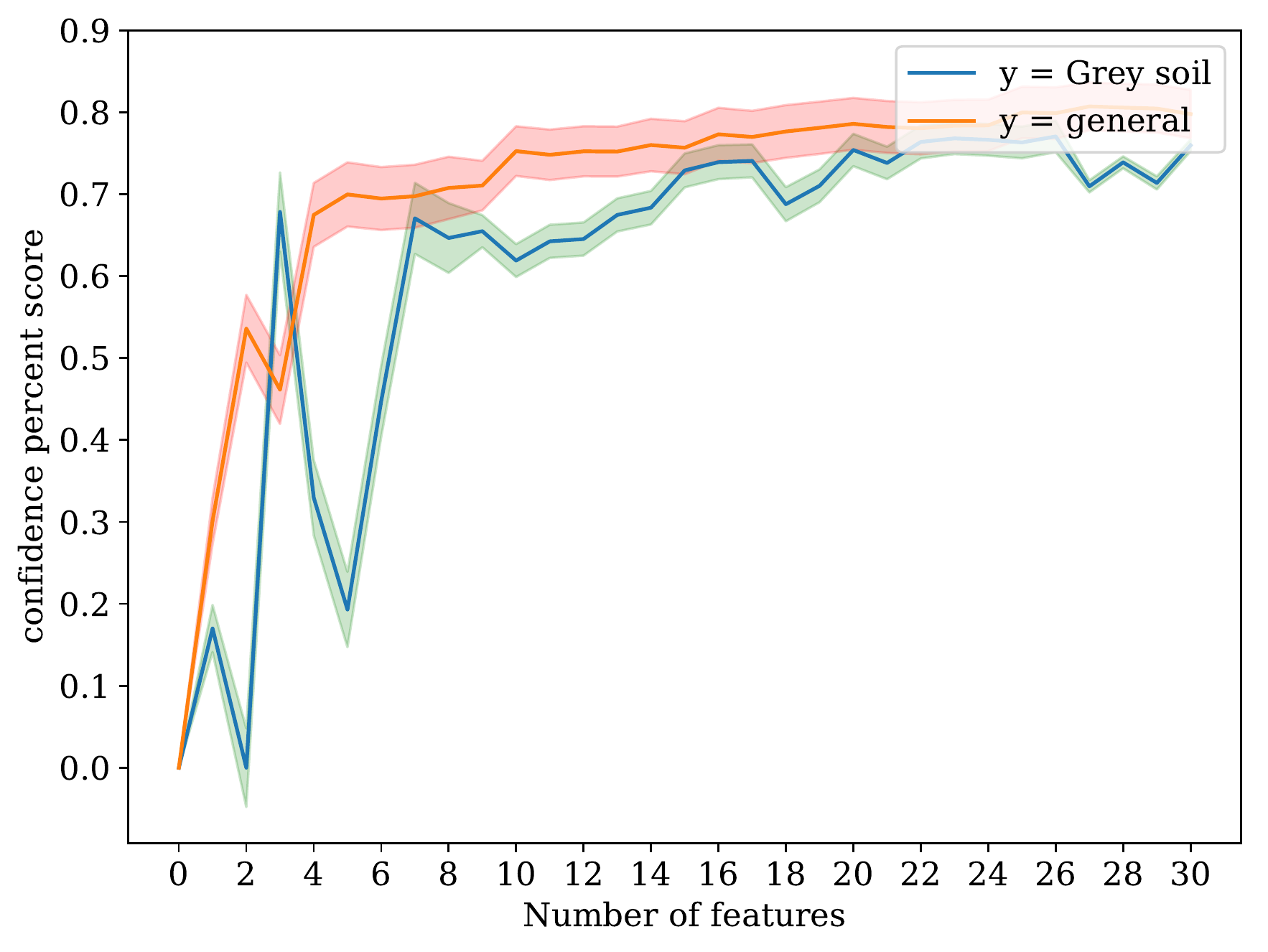}
        \label{fig:satlog_all}
        \caption{Non-target-specific confidence in for mRMR on the Satlog dataset}
    \end{minipage}
    ~
    \begin{minipage}[t]{0.3\textwidth}
	\centering	
        \includegraphics[width=\linewidth]{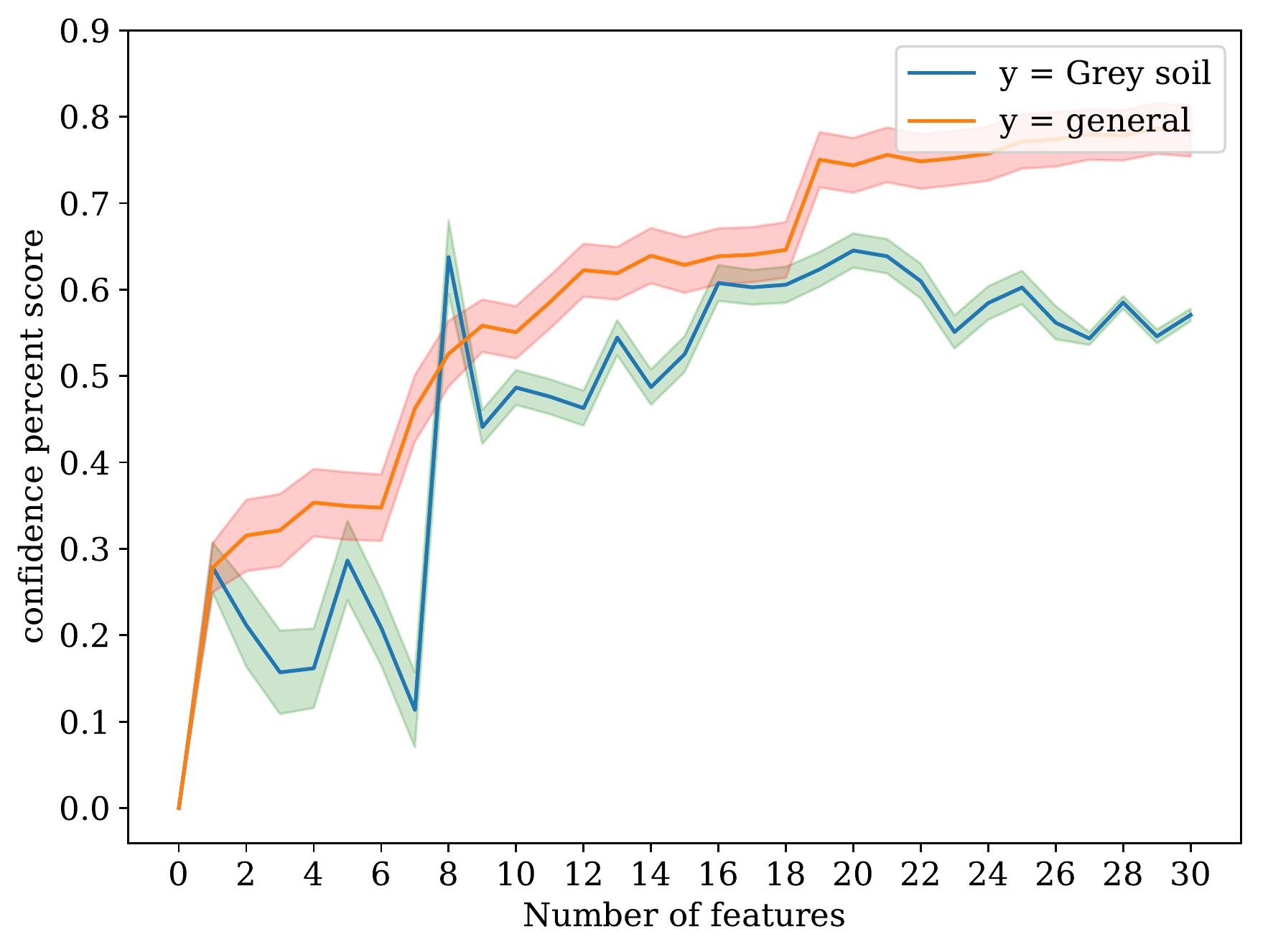}
        \label{fig:breast_cancer_all}
        \caption{Non-target-specific confidence in for Lasso on the Satlog dataset}
    \end{minipage}
    ~
    \begin{minipage}[t]{0.3\textwidth}
	\centering	
        \includegraphics[width=\linewidth]{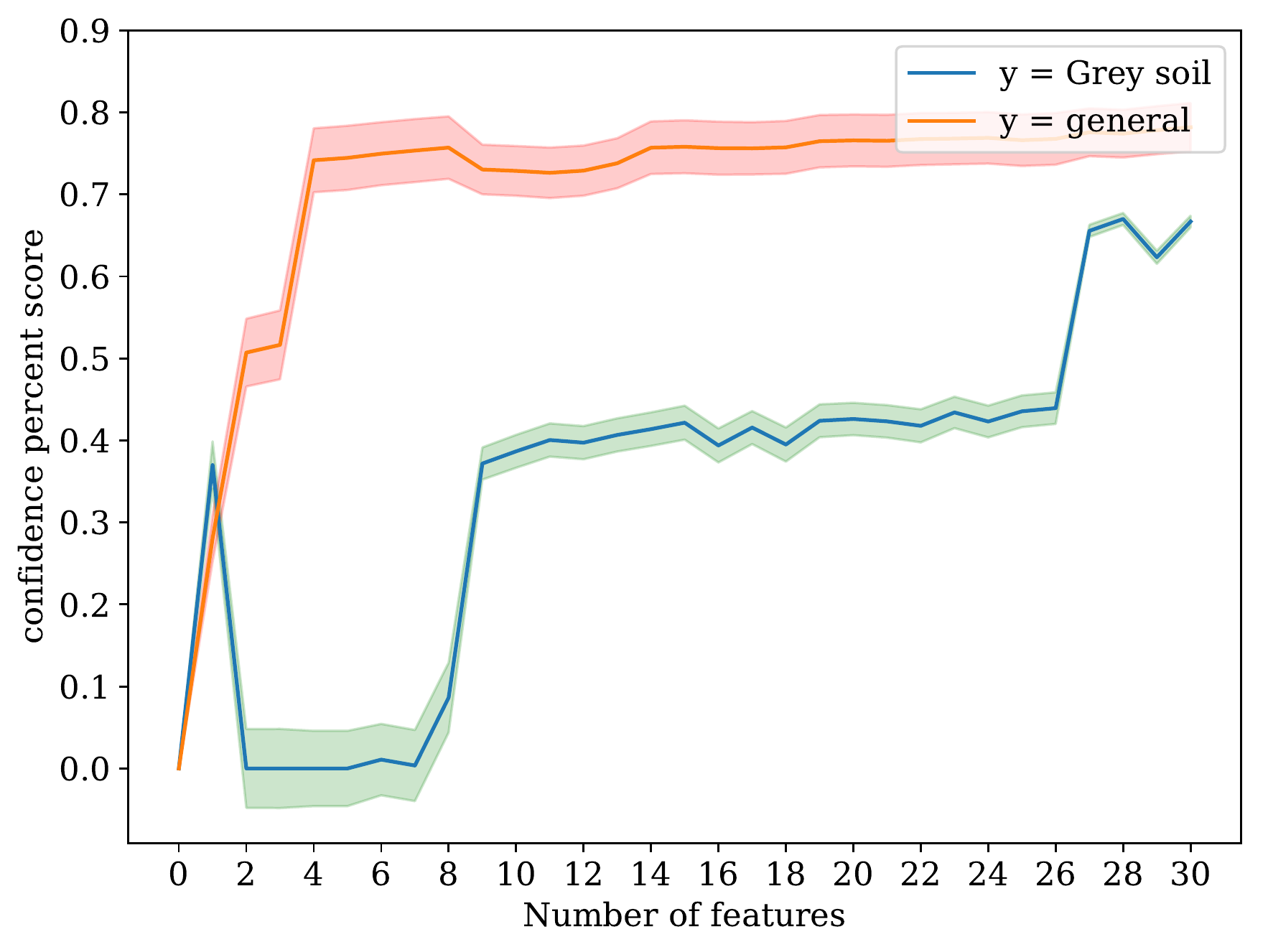}
        \label{fig:satlog_all}
        \caption{Non-target-specific confidence in for Extra Trees on the Satlog dataset}
    \end{minipage}
    ~
    \begin{minipage}[t]{0.3\textwidth}
	\centering	
        \includegraphics[width=\linewidth]{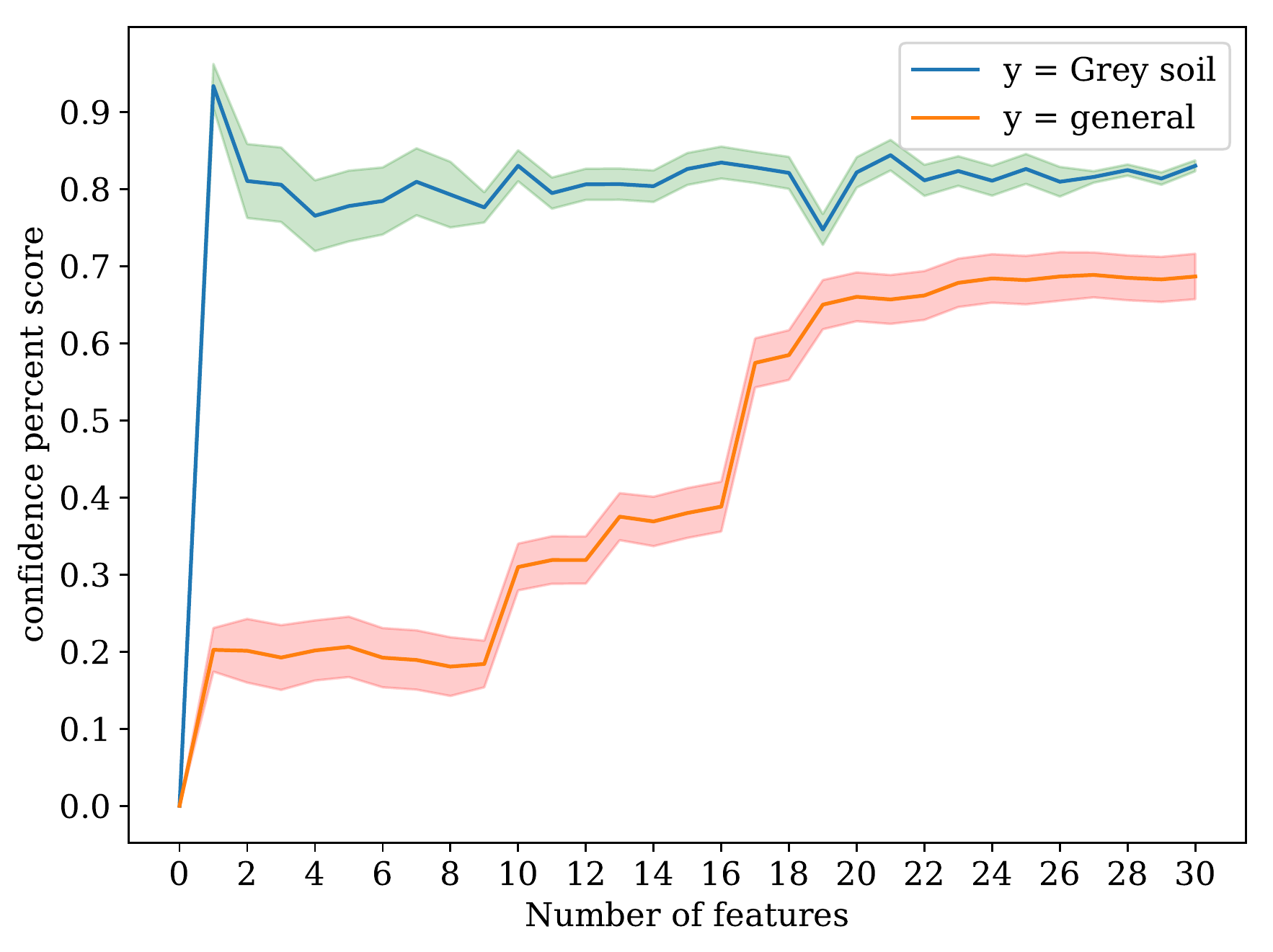}
        \label{fig:satlog_all}
        \caption{Non-target-specific confidence in for Target Focused on the Satlog dataset}
    \end{minipage}
\end{figure}

\subsubsection{NHANES Heart}
\begin{figure}[H]
    \centering
    \begin{minipage}[t]{0.3\textwidth}
	\centering	
        \includegraphics[width=\linewidth]{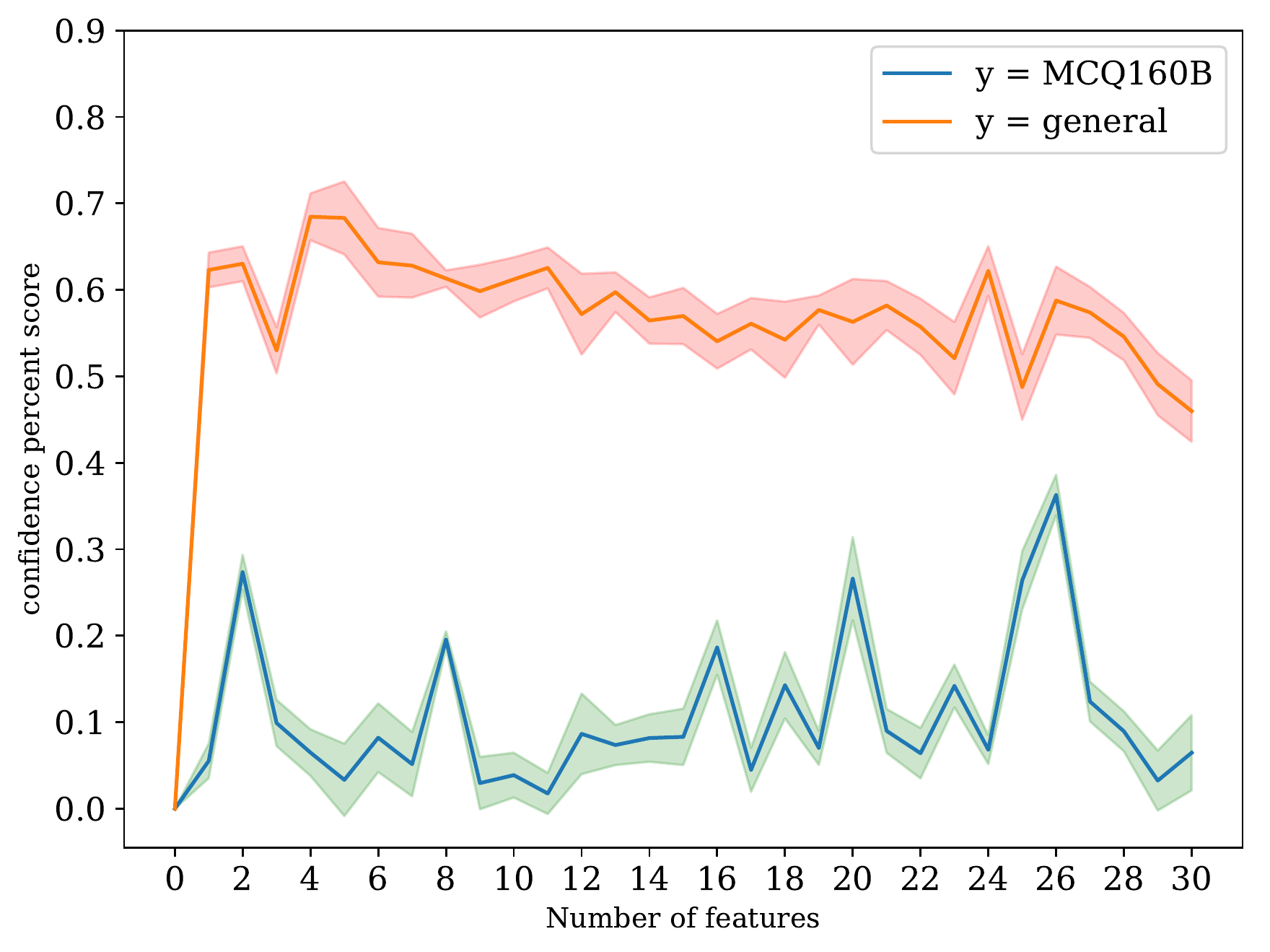}
        \label{fig:breast_cancer_all}
        \caption{Non-target-specific confidence in for mutual information on the NHANES heart dataset}
    \end{minipage}
    ~
    \begin{minipage}[t]{0.3\textwidth}
	\centering	
        \includegraphics[width=\linewidth]{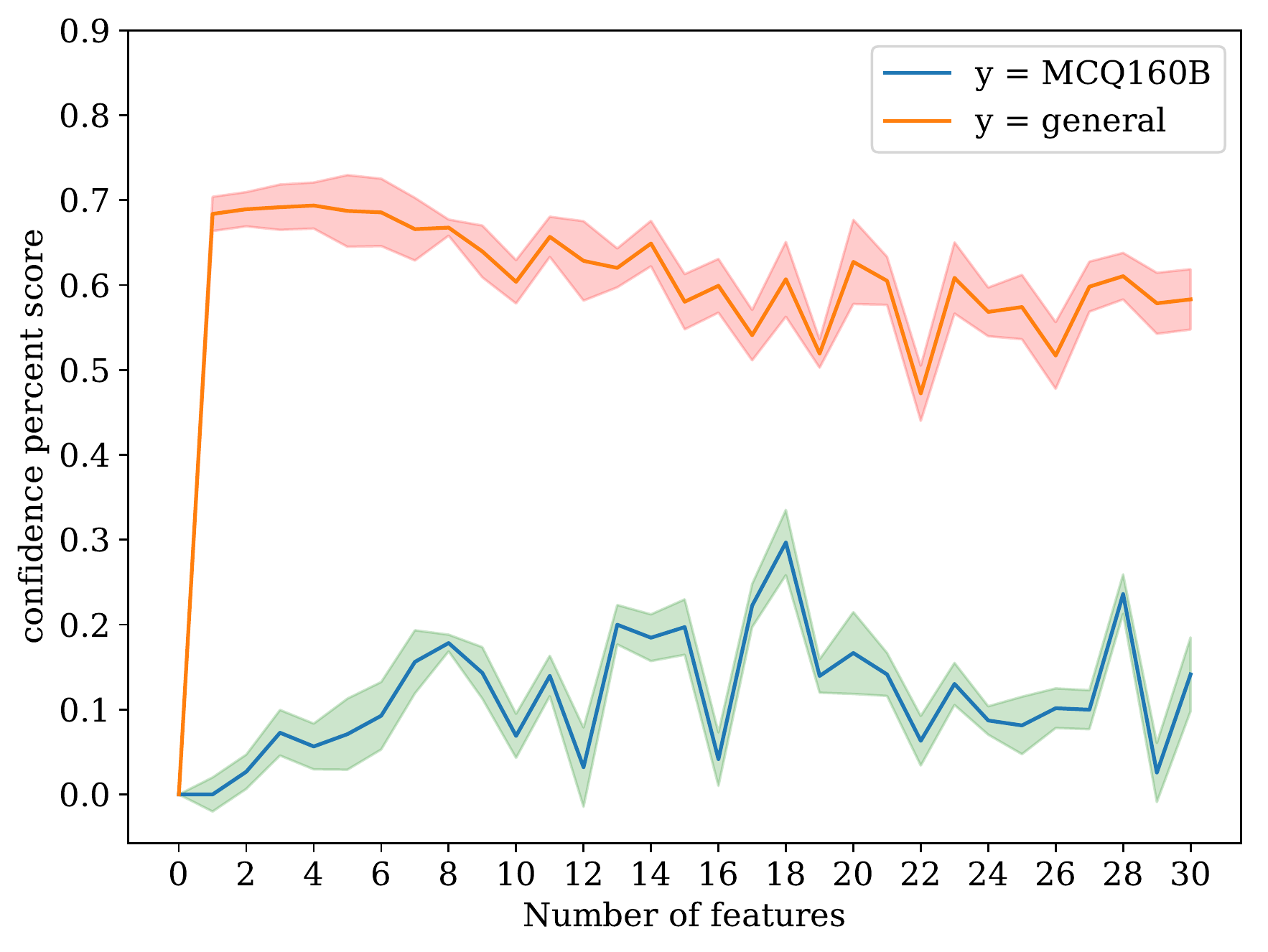}
        \label{fig:satlog_all}
        \caption{Non-target-specific confidence in for mRMR on the NHANES heart dataset}
    \end{minipage}
    ~
    \begin{minipage}[t]{0.3\textwidth}
	\centering	
        \includegraphics[width=\linewidth]{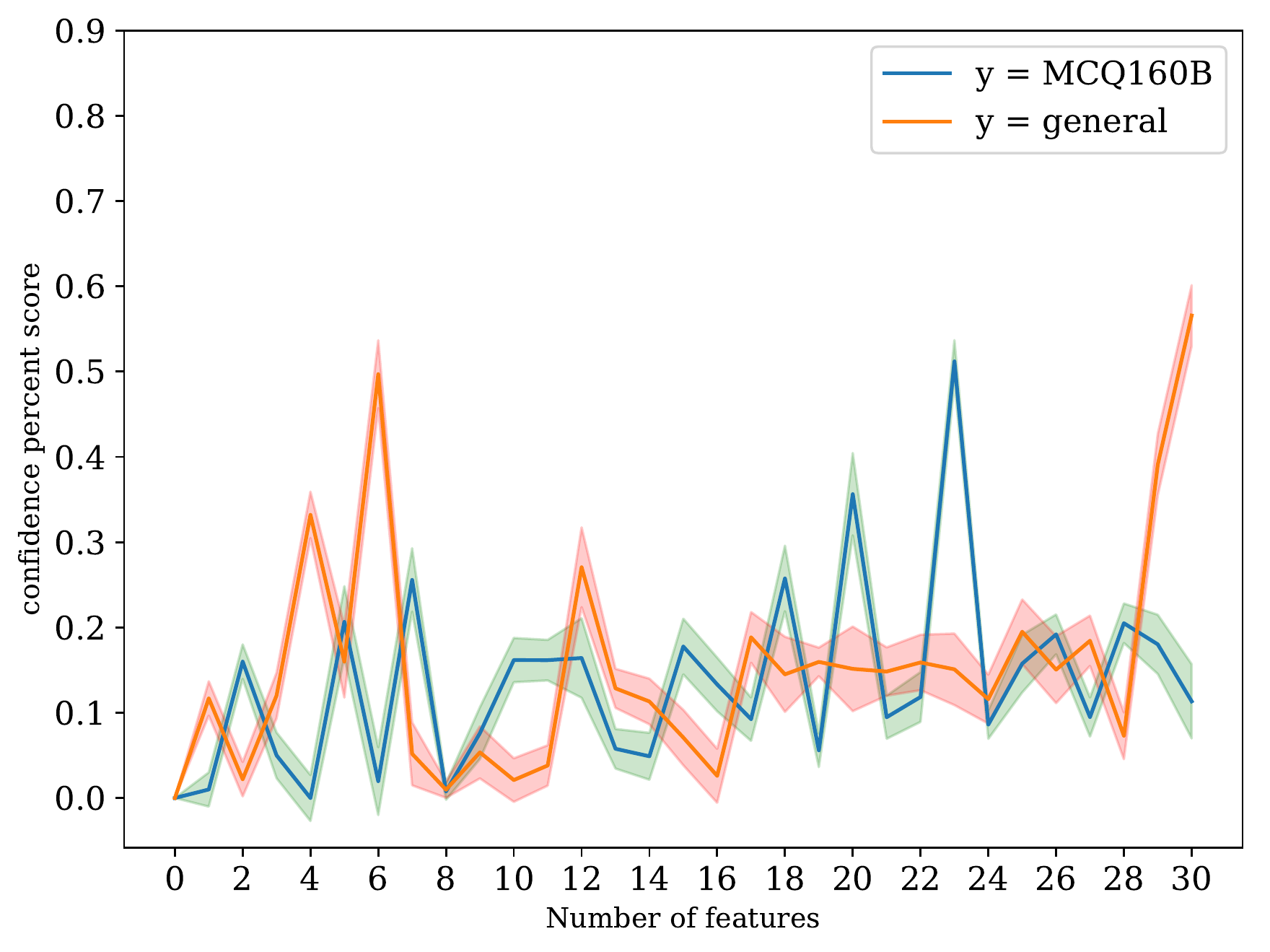}
        \label{fig:breast_cancer_all}
        \caption{Non-target-specific confidence in for Lasso on the NHANES heart dataset}
    \end{minipage}
    ~
    \begin{minipage}[t]{0.3\textwidth}
	\centering	
        \includegraphics[width=\linewidth]{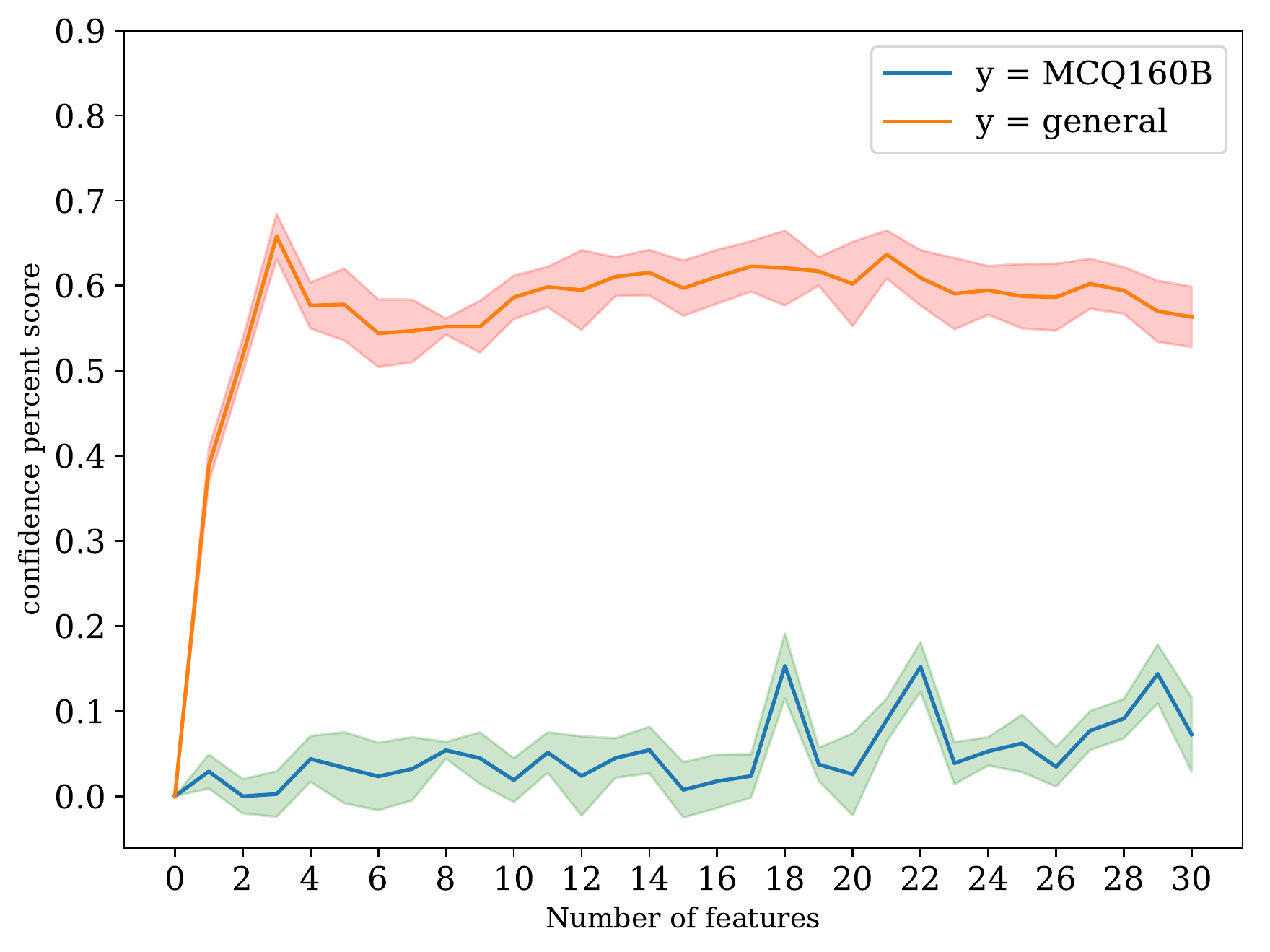}
        \label{fig:satlog_all}
        \caption{Non-target-specific confidence in for Extra Trees on the NHANES heart dataset}
    \end{minipage}
    ~
    \begin{minipage}[t]{0.3\textwidth}
	\centering	
        \includegraphics[width=\linewidth]{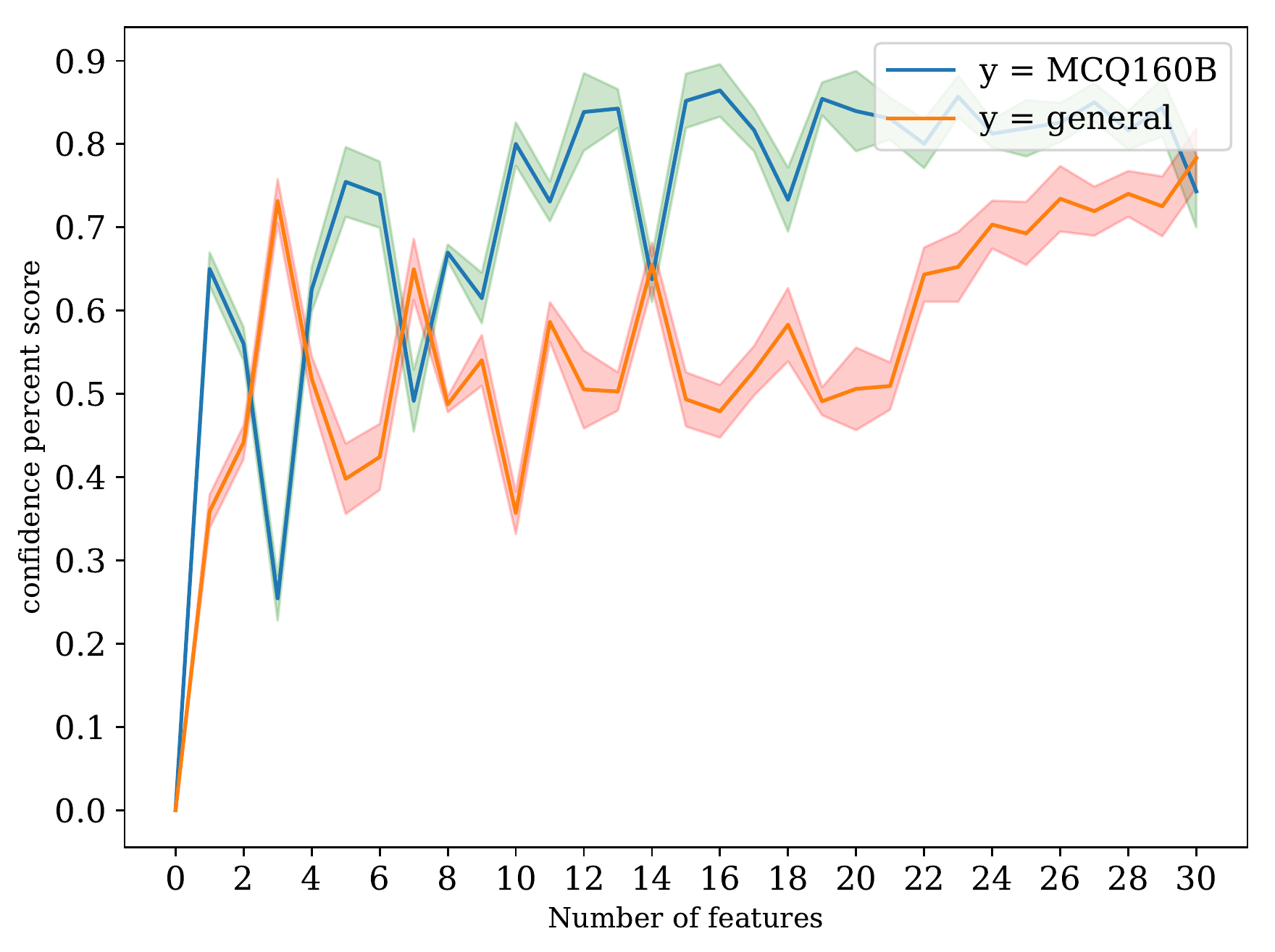}
        \label{fig:satlog_all}
        \caption{Non-target-specific confidence in for Target Focused on the NHANES heart dataset}
    \end{minipage}
\end{figure}

\subsubsection{NHANES Diabetes}
\begin{figure}[H]
    \centering
    \begin{minipage}[t]{0.3\textwidth}
	\centering	
        \includegraphics[width=\linewidth]{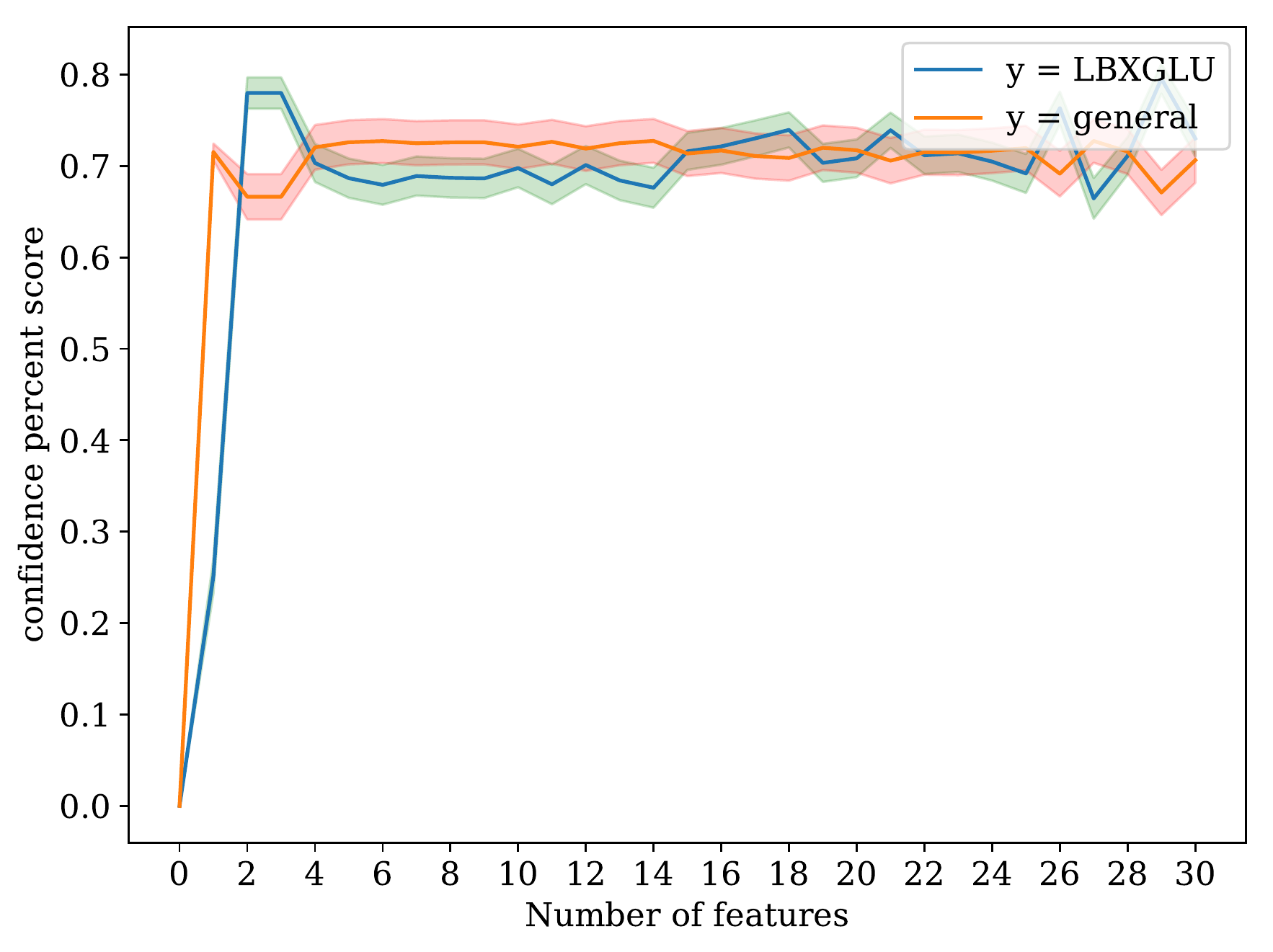}
        \label{fig:breast_cancer_all}
        \caption{Non-target-specific confidence in for mutual information on the NHANES diabetes dataset}
    \end{minipage}
    ~
    \begin{minipage}[t]{0.3\textwidth}
	\centering	
        \includegraphics[width=\linewidth]{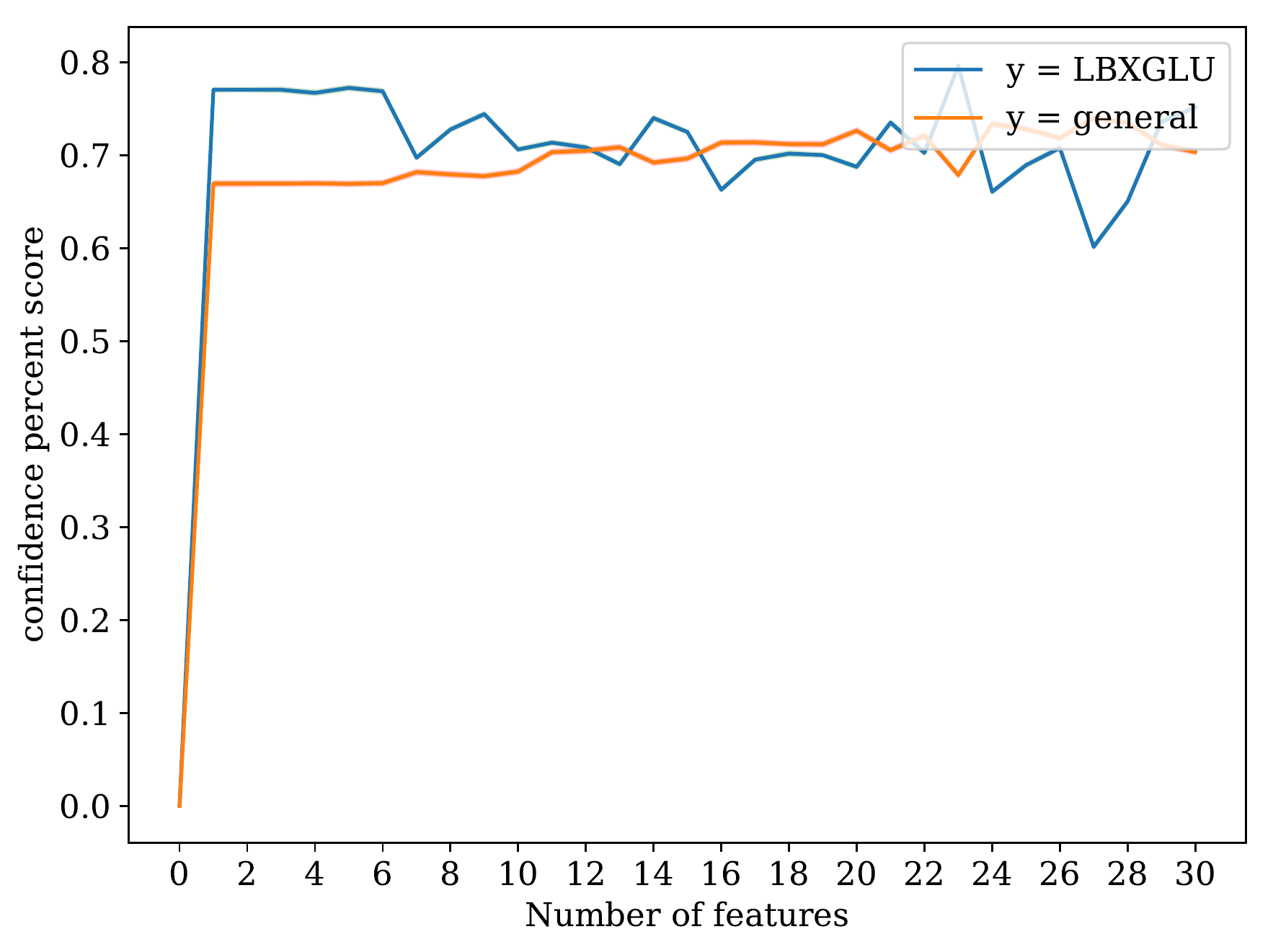}
        \label{fig:satlog_all}
        \caption{Non-target-specific confidence in for mRMR on the NHANES diabetes dataset}
    \end{minipage}
    ~
    \begin{minipage}[t]{0.3\textwidth}
	\centering	
        \includegraphics[width=\linewidth]{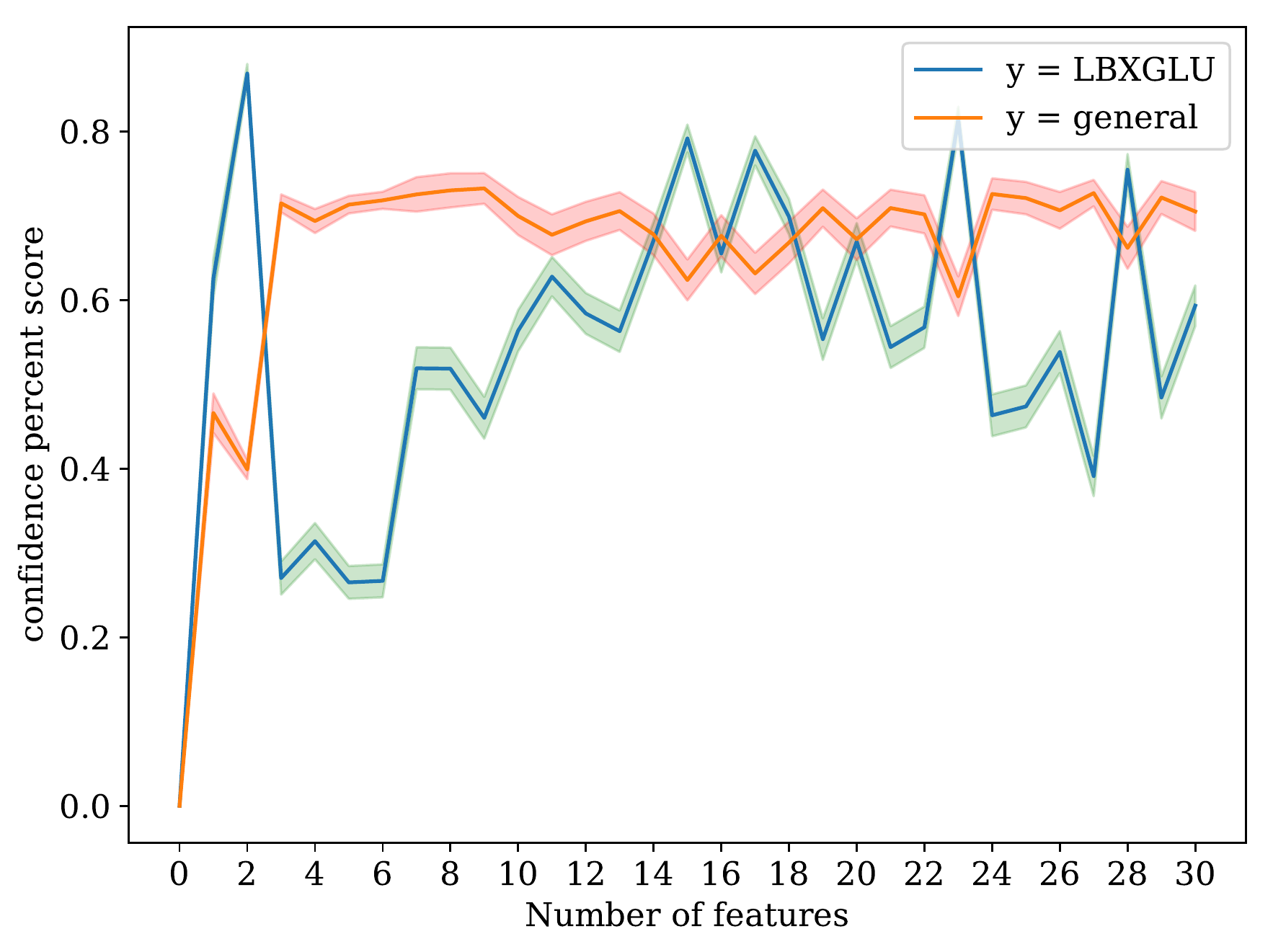}
        \label{fig:breast_cancer_all}
        \caption{Non-target-specific confidence in for Lasso on the NHANES diabetes dataset}
    \end{minipage}
    ~
    \begin{minipage}[t]{0.3\textwidth}
	\centering	
        \includegraphics[width=\linewidth]{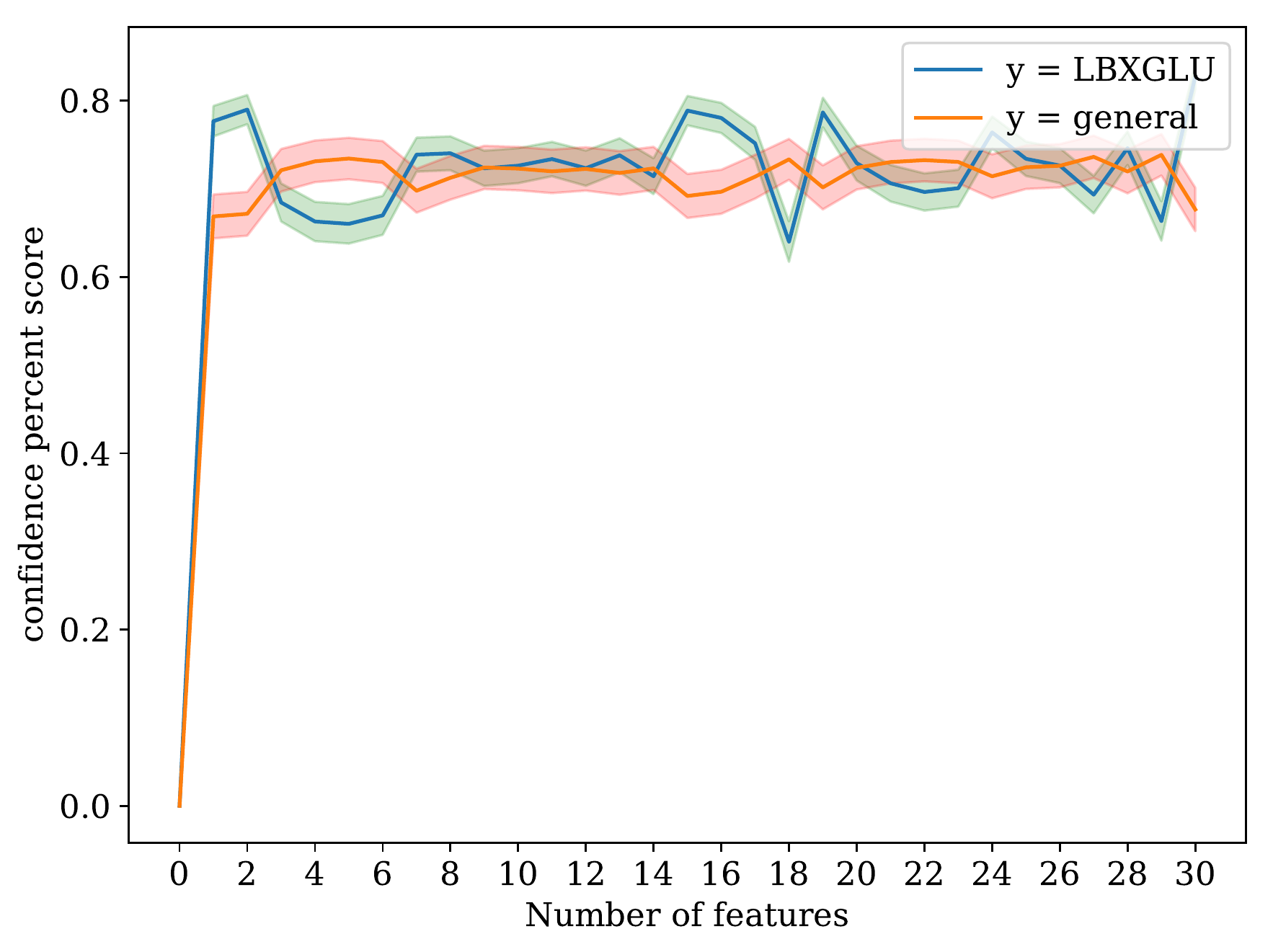}
        \label{fig:satlog_all}
        \caption{Non-target-specific confidence in for Extra Trees on the NHANES diabetes dataset}
    \end{minipage}
    ~
    \begin{minipage}[t]{0.3\textwidth}
	\centering	
        \includegraphics[width=\linewidth]{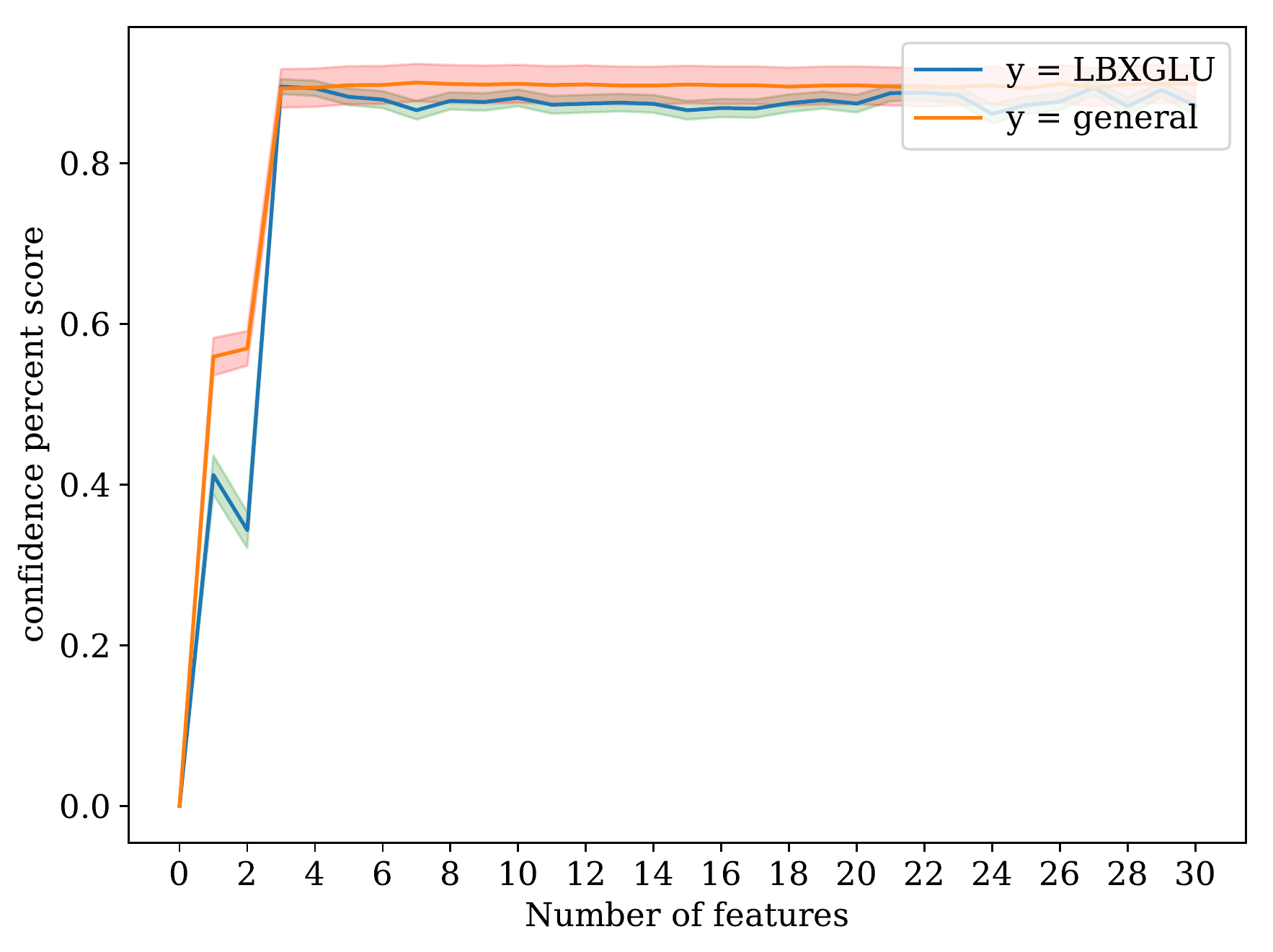}
        \label{fig:satlog_all}
        \caption{Non-target-specific confidence in for Target Focused on the NHANES diabetes dataset}
    \end{minipage}
\end{figure}

\section{Time Complexity Analysis}
Let $FS$ be the set collecting all selected features, $N$ be the number of available features, and assuming some constant budget $\beta$ for features. For a single feature $f_i \notin \textit{FS}$ we train a new model estimating the linear function $p(\mathbf{z}_n|\mathbf{X}_n^\top \mathbf{W}+\mathbf{b},\sigma^2_z)$ (equation 5 in the paper). The model is trained using a constant number of iterations and confidence is computed using a constant number of samples from the estimated distribution. $\textit{CovScore}$ and $\textit{CosScore}$ are both computed on the features already in $FS$ in time $2*\beta^2$. Since $\beta$ is constant, so is the time to compute $\textit{CovScore}$ and $\textit{CosScore}$. The final $v_i$ value is the product of another constant time multiplication. 

Once all features are scored, we append a single feature to the set $FS$, and the process starts again for $N-1$ features. Therefore, for $N$ features, the process will run $N + N-1 + N-2 ... + 1$ times. Resulting in an $N^2$ time complexity.